\documentclass[acmsmall]{acmart}

\AtBeginDocument{
  \providecommand\BibTeX{{
    \normalfont B\kern-0.5em{\scshape i\kern-0.25em b}\kern-0.8em\TeX}}}

\usepackage{subfigure}

\newcommand{\red}[1]{#1}

\begin{document}

\setcopyright{acmcopyright}
\acmJournal{TKDD}
\acmYear{2021} \acmVolume{1} \acmNumber{1} \acmArticle{1} \acmMonth{1} \acmPrice{15.00}\acmDOI{10.1145/3448021}

\title{Similarity Embedding Networks for Robust Human Activity Recognition}

\author{Chenglin Li}
\email{ch11@ualberta.ca}
\affiliation{%
  \institution{University of Alberta}
  \city{Edmonton, AB}
  \country{Canada}
  \postcode{T6G 2R3}
}

\author{Carrie Lu Tong}
\email{carrie.tong@ualberta.ca}
\affiliation{%
  \institution{University of Alberta}
  \city{Edmonton, AB}
  \country{Canada}
}

\author{Di Niu}
\email{dniu@ualberta.ca}
\affiliation{%
  \institution{University of Alberta}
  \city{Edmonton, AB}
  \country{Canada}
}

\author{Bei Jiang}
\email{bei1@ualberta.ca}
\affiliation{%
  \institution{University of Alberta}
  \city{Edmonton, AB}
  \country{Canada}
}

\author{Xiao Zuo}
\email{royzuo@tencent.com}
\affiliation{%
 \institution{Tencent}
 \city{Shenzhen}
 \country{China}
}

\author{Lei Cheng}
\email{raycheng@tencent.com}
\affiliation{%
 \institution{Tencent}
 \city{Shenzhen}
 \country{China}
}

\author{Jian Xiong}
\email{janexiong@tencent.com}
\affiliation{%
 \institution{Tencent}
 \city{Shenzhen}
 \country{China}
}

\author{Jianming Yang}
\email{kimmyyang@tencent.com}
\affiliation{%
  \institution{Tencent}
  \city{Shenzhen}
  \country{China}
}


\begin{abstract}
Deep learning models for human activity recognition (HAR) based on sensor data have been heavily studied recently. However, the generalization ability of deep models on complex real-world HAR data is limited by the availability of high-quality labeled activity data, which are hard to obtain.
In this paper, we design a similarity embedding neural network that maps input sensor signals onto real vectors through carefully designed convolutional and LSTM layers. The embedding network is trained with a pairwise similarity loss, encouraging the clustering of samples from the same class in the embedded real space, 
and can be effectively trained on a small dataset and even on a noisy dataset with mislabeled samples. 
Based on the learned embeddings, we further propose both nonparametric and parametric approaches for activity recognition. 
Extensive evaluation based on two public datasets has shown that the proposed similarity embedding network significantly outperforms state-of-the-art deep models on HAR classification tasks, is robust to mislabeled samples in the training set, and can also be used to effectively denoise a noisy dataset.
\end{abstract}





\maketitle
%  section 1: intro
\section{Introduction}
\label{sec:intro}
Human activity recognition (HAR) has become a fundamental capability in a wide range of Internet of Things (IoT) applications, including human health and well being \cite{alemdar2010wireless}, mobile security \cite{wang2016friend,miluzzo2012tapprints}, tracking and imaging \cite{li2015human} and vehicular road sensing \cite{kang2015ecodrive,hu2014towards}. The widespread deployment of sensor technology, especially motion sensors like the accelerometer and gyroscope on mobile phones and smart watches, has made mobile sensing ubiquitous and fuelled HAR research with data.

HAR is essentially a time series classification task based on a collection of heterogeneous measurements from multiple motion sensors. Traditional machine learning approaches such as decision trees, support vector machines \cite{svm12har}, hidden Markov models \cite{hiddenmc} and random forest \cite{comparison14} models have been developed based on manually extracted features and achieved decent performance in some controlled environments \cite{bulling2014tutorial}. However, these conventional pattern recognition algorithms heavily rely on the features correctly engineered from the raw data based on domain expertise and signal processing techniques, and therefore have limited generalizability. 

Recently, deep learning has achieved great advances in fields such as computer vision \cite{krizhevsky2012imagenet} and natural language processing \cite{mikolov2010recurrent,liu2018matching}. Popular deep learning models, including convolutional neural networks (CNNs), recurrent neural networks (RNNs), Restricted Boltzmann Machine (RBM) and their hybrids, have also been adopted for HAR tasks \cite{radu2016towards,bhattacharya2016smart} to overcome the limitations of manual feature engineering. For example, DeepSense \cite{yao2017deepsense}, as a deep model, has achieved the state-of-the-art HAR performance. 

These studies have fine-tuned the architecture of the deep neural networks and achieved good prediction accuracies in carefully controlled environments. 
For example, the USC-HAD dataset \cite{zhang2012usc} was collected from a special device, MotionNode, with a fixed sensing position (all users collect data by wearing MotionNode at the same position). Thus, even using CNNs with only 4 hidden layers, it can achieve a test accuracy of 97.01\% \cite{jiang2015human}. 
On the other hand, the HHAR (Heterogeneous Human Activity Recognition) dataset collected by \cite{stisen2015smart} is a more complex one, with heterogeneous sampling rates, hardware devices and mobile operating systems. On this dataset, even with data augmentation techniques such as adding random noise, interpolation, and resampling \cite{deepsense_pro}, 
the state-of-the-art model, DeepSense which combines RNN with multiple convolutional layers, can only achieve a 94\% accuracy. Therefore, more robust models need \red{to} be developed to handle increasingly complex and non-uniform sensor data from diverse real-world devices. 

Furthermore, to date, two important factors have hindered the widespread deployment of deep models in real-world HAR applications. \emph{First}, it is commonly known that deep neural networks need to be trained on a large amount of samples in order to yield competent performance. However, it is hard to collect a large amount of labeled samples to train HAR classifiers. Most HAR datasets reported in the literature contain no more than a few or a dozens of users (with the largest study based on 48 users \cite{wang2018deep}) and no more than tens of thousands of processed training samples, a size not comparable to the use of deep models in other fields, which usually have datasets of millions of samples or more, e.g., ImageNet.
\emph{Second}, in real world HAR applications, training data are usually collected from crowd sourcing and may be subject to labeling mistakes or noise due to careless or malicious users.
Furthermore, unlike images, speeches or text, once sensor signals are collected, it is extremely hard to authenticate the correctness of their activity labels. 
As a result, for HAR problems, usually either a high-quality yet small dataset or a larger dataset with label noise is available. This explains why existing deep neural networks, which are originally designed for large and clean datasets, do not yield close-to-perfect performance on HAR tasks.

To overcome the aforementioned issues, in this paper, we propose a robust \emph{similarity embedding network} (SEN) that can generate discriminative vectorized embeddings of input signals based on a small amount of training data. Such embeddings can directly be used for HAR, greatly boosting the classification accuracy, or for data denoising. We have made multiple contributions:

First, similar to DeepSense, we leverage both CNNs and RNNs to model the interaction among different sensor signals within a same time interval and the sequential dependencies of signals across time intervals. Yet, we represent input signals differently and apply convolutional and LSTM layers in a new way that is optimized for performance. We show that our network structure alone when directly used for activity recognition is comparable to other state-of-the-art deep models, even at a much lower sampling rate of sensor data.

More importantly, we introduce the similarity embedding network (SEN), which is based on the proposed neural network, yet trained by approximating the class label similarity between any two samples using the cosine similarity between their embeddings in a real space.   
This replaces the cross-entropy loss in the original classification problem by a pairwise loss between a pair of samples, such that the embeddings of samples from the same class are pushed to cluster, while those of different classes are pushed to separate from each other. 
Since every pair of samples yields a loss value to optimize the proposed similarity embedding network, we in fact have bootstrapped a small dataset into a larger amount of training samples, while building robustness into the model.  
We discuss the connections of our technique to as well as its differences from learning to hash \cite{wang2018survey}, matching networks \cite{vinyals2016matching} and one-shot learning \cite{koch2015siamese}.

Based on the signal embeddings trained with the proposed SEN, we further propose two methods that achieve state-of-the-art activity recognition performance, including 1) a nonparametric nearest neighbor approach applied onto the embedded vectors; and 2) a MLP classifier applied on top of the similarity embedding network. We demonstrate, through experimental results, that the nonparametric nearest neighbor approach achieves strong robustness in the presence of small training sets and label noise.

In addition, we also propose a simple denoise procedure based on the trained SEN, which can leverage a small clean set to detect mislabeled samples in a potentially contaminated dataset.

We demonstrate the superiority and robustness of the proposed similarity embedding network on two publicly available datasets, the \emph{HHAR} dataset and \emph{USC-HAD} dataset.  
The proposed classification methods based on the embeddings found by SEN significantly outperform the state of the art in deep neural HAR. 
On the augmented HHAR dataset (following the same data augmentation procedure as DeepSense \cite{deepsense_pro}), our proposed model architecture achieved an accuracy over 98\%, outperforms the 94.2\% accuracy reported by DeepSense in a leave-one-user-out evaluation scheme when trained with cross-entropy loss on HHAR dataset.  
During stress test on the original HHAR dataset, by only using 110 training samples per class (7\% of all data), our SEN-based method already achieves an accuracy of 95.03\%, which is the same accuracy achieved by traditional classification-oriented deep models which, however, use 80\% of data for training. 

Furthermore, we demonstrate the robustness and generalizability of the proposed SENs by training the model on datasets with different noise rates. Results show that the proposed SEN-SM method is much more robust to labeling noise compared to classification-oriented deep neural networks.  
Finally, we also demonstrate the usefulness of the proposed SEN in terms of denoising a heavily contaminated dataset (with 40\% mislabeled samples) using embeddings trained on a small amount of clean data. 

\begin{figure*}. % show 
  \centering
  \includegraphics[width=5in]{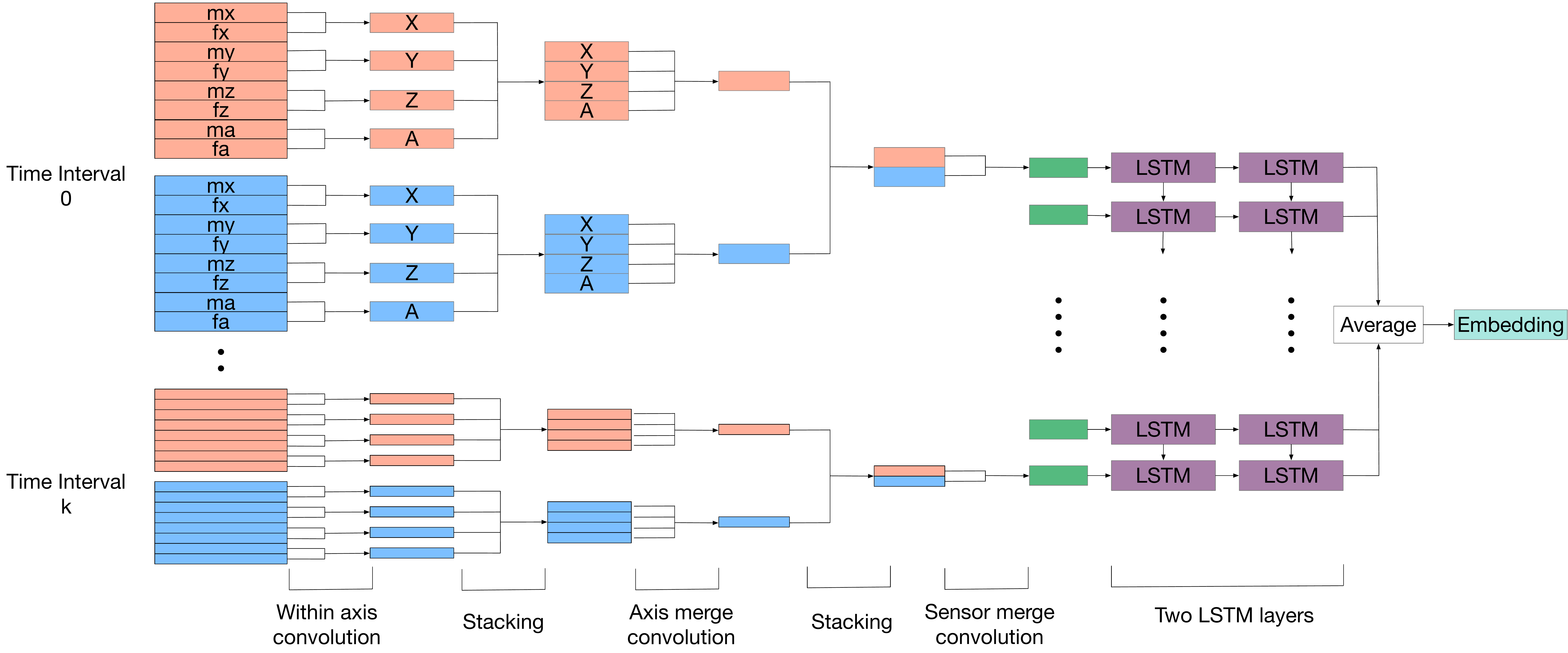}
  \caption{The proposed embedding network architecture. Boxes in orange represent data from accelerometer, while boxes in blue represent data from gyroscope. $mx, my, mz, ma$ denote the magnitudes after Fourier transforms and $fx, fy, fz, fa$ denote the corresponding frequencies after Fourier transforms.}
  \label{fig:structure}
  \Description[Architecture of proposed model]{The overall architecture of proposed model, where signal are proposed from the left to the right. }
\end{figure*}

% remaind of this paper{}
The remainder of this paper is organized as follows. We first introduce the proposed network architecture, training based on pairwise losses, and our HAR classification methods in Sec.~\ref{sec:model}. Extensive evaluation of the proposed methods is then performed on two datasets and presented in Sec.~\ref{sec:exp}. Sec.~\ref{sec:related} discusses the related work. Finally, we conclude the paper in Sec.~\ref{sec:conclude}. 

% Section 2  Model.
\section{Similarity Embedding Network}
\label{sec:model} 
In this section, we introduce the architecture of the similarity embedding neural network (SEN), pairwise loss that we adopted to train the SEN model. Approaches that leverage the embeddings output of SEN for HAR task are also discussed in this section. 
\subsection{Sensor Signals Processing}

In this paper, an original data sample consists of readings from two motion sensors $S = \{S_1, S_2\}$, typically including the accelerometer $S_1$ and gyroscope $S_2$. Signals generated by each sensor $S_i$ form a matrix $D_i$ of the shape $d^i \times n^i$, where $d^i$ is the dimensionality of the sensor $S_i$ and $n^i$ is the length of the signal. Typically, $d^1 = d^2 = 3$ for motion sensors.
Our procedure to process input motion sensor signals is similar to yet different from other typical deep models like DeepSense. The differences will be explained at the end of this subsection.

Considering one particular sensor $S_i$ as an example. 
Let $x_i, y_i, z_i$ be the time series data of sensor $S_i$ along the $x,y,z$ axes, respectively.
We first compute the amplitude series as $a_i = \sqrt{x_i^2 + y_i^2 + z_i^2}$. The amplitude series $a_i$ is then added to the original data of sensor $S_i$ as its $(d^i + 1)$th axis.
Thus, the data of sensor $S_i$ becomes a matrix of shape $(d^i + 1) \times n^i$. 

In order to leverage RNNs to model sequential dependencies, we split the data into $k$ time intervals, each of width $\tau$, which is a tunable hyper-parameter. 
The Fourier transform is then applied to the time series of each time interval in each of the $d^i + 1$ axes to get its frequency domain representation, which in our model is represented by both a vector of magnitudes and their corresponding frequencies. 
For example, the input series from axis $x$ of sensor $S_i$ is transformed into two vectors $mx$ and $fx$, representing the magnitudes and their corresponding frequencies after the Fourier transform, respectively. 
Then, we stack all the outputs from Fourier transforms into a $2(d ^ i + 1) \times f$ matrix, where $f$ is the length of frequency domain representations. 
We apply the same procedure to all time intervals of sensor $S_i$ and then stack them into a $k \times 2(d^i + 1) \times f $ tensor $X^i$. 
Finally, the set of resulting tensors for each sensor, $\mathcal{X} = \{X^i\}, i =1,2$, will be the input into our embedding network.

Note that our signal processing method differs from typical deep models for HAR like DeepSense in two ways: 1) we introduce the amplitude series $a_i$; 2) for each time interval at each sensor, we stack data into a $2(d^i + 1) \times f$ matrix with each of the $d^i +1$ axes expanded into two rows (including magnitudes and frequencies after the Fourier transform), whereas DeepSense processes the same interval of data into a  long $d^i \times 2f$ matrix using concatenation. 

\subsection{Network Architecture}
Our embedding network encodes the input of any mobile sensor reading sample, which is a $k \times |S| \times 2(d+1) \times f $ tensor $\mathcal{X}$ computed using the signal processing procedure mentioned above, into an embedded vector. 
As is shown in Fig. \ref{fig:structure}, the embedding network leverages both CNNs and RNNs in its neural encoding operation. 
Although prior research like DeepSense also adopted CNNs and RNNs to encode the time series from sensors, our network structure distinguishes from previous studies in multiple aspects. 

First, as has been mentioned above, the input of our network is a tensor of shape $k \times |S| \times 2(d+1) \times f$. As is shown in the left side of Fig. \ref{fig:structure}, in each interval, $mx$, $my$, $mz$, $ma$ represent the magnitudes (in frequency domain) and $fx$, $fy$, $fz$, $fa$ represent the corresponding frequencies. However, in DeepSense the input data of each sensor form a tensor of shape $d \times 2f \times k$. That is, for each axis, DeepSense merges the output of the Fourier transform into a single vector, whereas we treat the magnitude-frequency pair as two stacked vectors. By stacking vertically, we can apply convolutional layers and stacking layers recursively and hierarchically for better feature extraction. 

Second, similar to DeepSense, the data from each time interval is passed through a series of convolutional layers. But we apply the convolutional layers in a different way. DeepSense first applies three 1D convolutional layers to the $d\times 2f$ input from each sensor to get a vector and then, after stacking the vectors from different sensors, applies three other 1D convolutional layers to learn the interactions between sensors. 

In contrast, we apply \emph{hierarchical} convolutional schemes to the input of size $2(d+1) \times f$ from each sensor. As is shown in Fig. \ref{fig:structure}, first, we apply the \emph{within-axis convolution} with a filter of size $2\times conv1$ to the stacked pair of magnitudes and frequencies in each axis. Then we stack data from the four axes $x,y,z,a$ together and apply the \emph{axis-merge convolution} with a filter of size $4\times conv2$ to learn the interactions among axes. Finally, we stack the produced vectors from different sensors and apply the \emph{sensor merge convolution}, which consists of two convolutional layers with a $2\times conv3$ filter and a $1\times conv4$ filter, respectively. All the convolutions in our network are $1D$ convolutions, with filters scanning along one dimension horizontally. In contrast to DeepSense which has 6 convolutional layers, we only have 4 convolutional layers. By using hierarchical convolutions and within-axis convolutions, and by applying stacking repeatedly, we avoid running into a 1D vector too early.

% in chronological order
After the data of all $k$ time intervals have individually gone through the above hierarchical convolutional layers, we obtain $k$ CNN outputs in the order of time intervals. An RNN is then applied to these CNN outputs to capture sequential dependencies. A stacked structure with two layers of GRU is applied in DeepSense. Instead, we use two Long Short-Term Memory (LSTM) layers in our model which leads to better performance according to experiments. 
Finally, the $k$ outputs of the second LSTM are averaged to yield the final output of our network, which serves as the embedding of the input signal sample.

\subsection{Training with Pairwise Losses} 
\label{sec:pairloss} 
A key to the proposed similarity embedding network is to train it using pairwise losses instead of cross-entropy in classification. This will encourage the physical clustering of sample embeddings in a real space, bootstrap the training set and make the model robust to noise.
Given $N$ training samples $X = \{x_i\}_{i=1}^{N} \in \mathbb{R}^{m \times N}$, where $m$ is the input feature size, and their labels $Y = \{y_i\}_{i = 1}^{N}\in\mathbb{R}^{c \times N}$, where $y_i$ is a one-hot vector of length $c$, with $c$ being the number of activities, our objective is to learn an embedding function $h(\cdot)$ represented by the neural network presented above so that the embeddings of all samples $\hat{X} = \{h(x_i)\}_{i = 1}^{N} \in \mathbb{R}^{l \times N}$ are well clustered in the real space $\mathbb{R}^l$ according to their class labels. 

Let $E=\{e_i\}_{i=1}^N$, where $e_i = h(x_i)$, represent the set of all the embedded vectors. 
Let $s_{ij}$ denote the similarity between sample $i$ and sample $j$, with
$s_{ij} = 0$ if samples $i$ and $j$ are not from the same class, otherwise $s_{ij} = 1$. With the label information in the training set, the pairwise similarity relationships $S = \{s_{ij}\}$ can be easily derived. 
The Maximum a Posterior (MAP) estimation of the embeddings $E = {e_i}, i = 1,2,\ldots, N$ can be represented as ~\eqref{eq:map} 
\begin{equation} \label{eq:map}
p(E|S) \propto p(S|E)p(E) = \prod_{s_{ij} \in S} p(s_{ij}|e_i, e_j) p(e_i, e_j)
\end{equation}
where $p(S|E)$ denotes the likelihood function. For each pair of the input data, $p(s_{ij} | E)$ is the conditional probability of $s_{ij}$ given the embeddings $E$, which is defined as ~\eqref{eq:cond_p}
\begin{equation} \label{eq:cond_p}
p(s_{ij}|E) = \lbrace_{1 - \sigma({\Phi_{ij}}),\quad s_{ij} = 0} ^{\sigma({\Phi_{ij}}), \quad s_{ij} = 1}
\end{equation}
where $\sigma(x) = 1/(1+ e^{-kx})$ is the logistic sigmoid function with a tunable parameter $k$, and $\Phi_{ij}$ is the cosine similarity between embedding $e_i$ and $e_j$ defined in ~\eqref{eq:cosine_s}.
\begin{equation} \label{eq:cosine_s}
\Phi_{ij} = \frac{e_i^T e_j}{|e_i| \cdot |e_j|}
\end{equation} 

\begin{figure}. % show 
  \centering
  \includegraphics[width=2.2in,height = 1.5in]{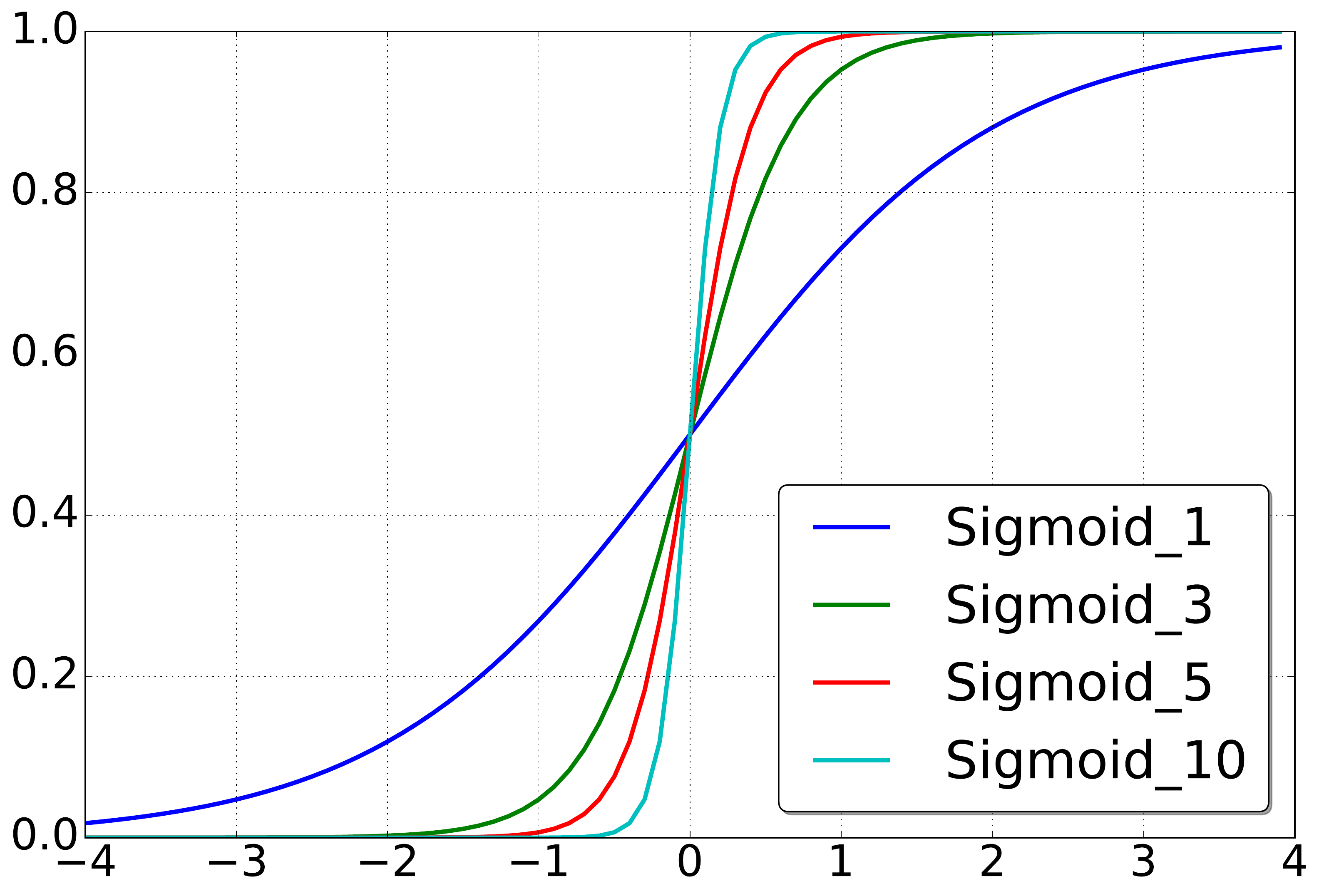}
  \caption{Logistic sigmoid function with different $k$ values.}
  \label{fig:sigmoid}
  \Description[Sigmoid Function]{Curves of Sigmoid functions with different $k$ values, when $k$ is bigger curve become steeper.}
\end{figure}

Obviously, if the conditional probability $p(s_{ij} | e_i, e_j)$ is maximized, the cosine similarity $\Phi_{ij}$ between the embedded vectors $e_i$ and $e_j$ will reach the maximum if they are from the same class and $\Phi_{ij}$ will reach the minimum if they are from different classes. 
It is worth noting that the value of the cosine similarity falls in the range of $[-1, 1]$, which is too narrow a section for the standard logistic sigmoid function where $k = 1$. To get better distinguish capability of embedding pairs with different cosine similarities, in our model, we set $k$ to be 10 so that we can almost cover the whole probability range from 0 to 1. As illustrated in Fig.~\ref{fig:sigmoid} when $k=1$ the covered probability range is approximately from 0.3 to 0.7 while is almost 0 to 1.0 when $k=10$. 

To maximize $p(S|E)$, the negative log likelihood function $J$ defined in~\eqref{eq:loss} is used to learn the embeddings. Minimizing $J$ by choosing the optimal embeddings networks that generate $E = {e_i}, i= 1,2, \ldots, N$ will make the cosine similarity of any two samples from the same class as large as possible, while in the mean time pushing the similarities between samples of different classes to as small as possible.

\begin{equation} \label{eq:loss}
  \displaystyle
  J = -\sum_{s_{ij} \in S}log\ p(s_{ij}|e_i, e_j) = -\sum_{s_{ij} \in S}(s_{ij} k \Phi_{ij} - log\ (1 + e^{ k \Phi_{ij}}))
\end{equation}

To minimize $J$, we can sample a batch of $B$ pairwise samples, each in the form of $(x_i, x_j, s_{ij})$ to perform batched learning. For each training sample $(x_i, x_j, s_{ij})$, $e_i=h(x_i)$ and $e_j=h(x_j)$ are obtained from the same embedding network $h(\cdot)$. And the parameters of $h(\cdot)$ will be updated by error back-propagation according to ~\eqref{eq:loss}. 
Popular optimization algorithms such as stochastic gradient descent (SGD), Adam, and AdaGrad can be adopted to minimize $J$. 

There are many studies in the field of learning to hash \cite{wang2018survey,li2017deep} (mostly applicable to images) that also minimize pairwise losses. However, they are different from our method in the following aspects. First, studies in learning to hash focus on similarity preserving from the original feature space to the hash space. The similarity in the original space is often measured by $l_2$-norm or cosine distance of raw features. While in our model, the original similarity to be preserved is whether two samples belong to the same class. Second, the embeddings in learning to hash are binary vectors, where the hamming distance is used to measure the similarity between two embedded vectors mainly to speedup retrieval. On the other hand, the embeddings from our network are real vectors such that the cosine similarity can be used to maximize the separation of across-class samples. Third, the problem settings of learning to hash are also different: they aim to speed up information retrieval, especially of image objects, and do not suffer from a lack of training data or noisy data with mis-labeled samples.

Our work is also related to matching networks \cite{vinyals2016matching} and one-shot learning \cite{koch2015siamese}, which perform image recognition based on a small number of samples.
Matching networks \cite{vinyals2016matching} still use cross-entropy loss between predicted labels and the ground truth labels to train the network while we use the pairwise similarities between samples to learn embeddings. The prediction of a new sample relies on the selected support set. 
One-shot learning uses pairwise losses in a Siamese network. However, it minimizes a cross-entropy loss and uses the $l1$-distance measure of almost binary embeddings (similar to hamming distance), while we maximize the log likelihood of observing the similarities given embeddings and embed vectors in a real space to maximize separability.

% Section 3 classification
\subsection{Activity Recognition}
\label{sec:application}
To leverage the proposed SENs for HAR task, we propose two algorithms which take the embeddings of samples calculated by a pretrained similarity embedding network as inputs. By minimizing the pairwise loss defined in ~\eqref{sec:pairloss}, the embeddings of all the samples are expected to be well clustered. Hence, the most straightforward way to predict the label of a newly arrived sample is through the nonparametric k-nearest neighbors (k-NN) algorithm. To speed up prediction and also enhance robustness to noise, we propose to perform similarity matching between the new sample and the class centers in the embedded space. That is, the label for a new sample is predicted to be the label of the class center that is the closest to the new sample in the embedded space. The class center is calculated by averaging the normalized embeddings of all training samples in that class, with the distance measure being cosine distance. If the label noise is randomly scattered in the embedded space, its impact to each class center is reduced. 
Thus we argue that this SNEs with similarity matching (SEN-SM) method for HAR is robust to label noise and can achieve high accuracy even it is trained on noise dataset.   
We will further show experimental results in the following section to show the robustness of SEN-SM method. 

In addition, we also propose to use another multilayer perceptron (MLP) model with one hidden layer, to predict a sample's label only based on its embedding vector without using training data in prediction. We call this HAR method SEN-MLP. Here the MLP layer is trained \emph{separately} from the embedding network on the same training dataset. 
With the final MLP layer, the whole model structure looks similar to a typical deep model for classification. However, they are essentially different. In our model the embedding network and the MLP layer are trained separately, with sampled pairwise losses and the cross-entropy loss, respectively, whereas in typical classification deep models, they are jointly trained only with the cross-entropy loss, which does not emphasize the separability or clustering of the embeddings. As the MLP layer is trained separately with the same training data, the SEN-MLP algorithm does not have noise robustness.

The SEN-MLP method have both advantages and disadvantages over the SEN-SM approaches. To begin with, as a nonparametric method the SEN-SM is simple, fast and is robust to label noise training samples. However, the calculation of class centers is hard when the dataset is huge, causing the updating of class centers time consuming, thus the SEN-SM model is not suitable for on-line learning applications where model updating is common. For SEN-MLP method, even though the last MLP layer need to be trained separated with the SEN model and its not noise robust, it is much convenient to update the weights of the last MLP layer compared to the updating of class centers in the on-line learning settings.

Mentioned above, if the MLP layer and SENs are trained \emph{jointly} with the cross-entropy end-to-end manner, instead of separately training with pairwise loss and cross-entropy loss, respectively, we get the Baseline method. The last MLP layer or fully connected layer projects the embeddings to class labels as shown in ~\eqref{eq:last} where $e_i$ is the embedding of sample $i$ from SENs part, $y'_i$ is the predicted label, $\sigma(\cdot)$ is the softmax function, $g(\cdot)$ is the activation function with $b$ and $b_0$ being the bias. 
\begin{equation} \label{eq:last}
y'_i = \sigma(b_0 + g(W \cdot e_i + b)) 
\end{equation} 
This Baseline algorithm is similar to previous studies such as Deepsense and the DCNN model in \cite{jiang2015human}. These models are all trained with an end-to-end fashion by minimizing the averaged cross entropy loss on training dataset (for classification), shown in ~\eqref{eq:cross_entropy} where $y_i$ is the true label of sample $i$, $c$ is the number of classes and $N$ is the number of samples in the training set. 
\begin{equation} \label{eq:cross_entropy}
loss = \frac{1}{N}\sum_{i}^{N} \sum_{j}^{c}-y_{ij}log(y_{ij}')
\end{equation} 
We will show the superiority of our baseline method with proposed neural network architecture compared to previous work such as DeepSense with leave-one-user-out results.  

% Section 4 experiment
\section{Experiments}
\label{sec:exp} 
In this section, experimental results on two public datasets are presented to show the efficiency and robustness of the proposed Similarity Embedding Networks for human activity recognition. 

\subsection{Datasets and Experiment Setup}

\begin{table}[tbp]
\caption{Details of the original and processed Public Datasets}
\begin{center}
\begin{tabular}{lccccc}  
% \toprule
\hline
Dataset     & \#Activity  & \#Users  & S.Rate    & Position  & \#Samples \\
% \midrule
\cline{2-6}
USC-HAD     & 12      & 14         & 100Hz     &  Hip    &    - \\
HHAR      & 6       & 9      & 50-250Hz  &  Waist    &    - \\
\cline{1-6}
Processed data  & \#Activity  & \#Users  & S.Rate    & Position  & \#Samples\\
\cline{2-6}
USC-HAD     & 6       & 14         & 25Hz      &  Hip    &  3,170 \\
HHAR          & 6       & 9      & 25Hz      &  Waist  &  9,335 \\
% \bottomrule
\hline
\end{tabular}
% \caption{Results of stress test using SEN-SM algorithm for activity recognition. The first column shows the number of training samples in each class.}
\label{tab:datasets_des}
\end{center}
\end{table} 

There are many publicly available datasets that are widely adopted by recent studies on HAR \cite{hammerla2016deep,ravi2016deep,wang2018deep}. In this paper, we evaluate our proposed and baseline methods on two public datasets reported by \cite{zhang2012usc} and \cite{stisen2015smart}, respectively. The detailed information of these two datasets are shown in Table \ref{tab:datasets_des}.  
The \emph{USC-HAD} \cite{zhang2012usc} dataset was collected in 2012 by MotionNode, a wired sensing platform, with a sampling rate of 100Hz, while the \emph{HHAR} \cite{stisen2015smart} dataset was collected in 2015 with 8 different types of smart phones with sampling rates varying from 50Hz to 200Hz. 

% Standing, Sitting, Walking, Upstairs, Downstairs and Running (USC)
% Standing, Sitting, Walking, Upstairs, Downstairs and Biking (HHARs)
Preprocessing is applied to maintain consistency between the two datasets and reformat the original data to fit our proposed architecture. We first select 6 types of activities, including ``Standing'', ``Sitting'', ``Walking'', ``Upstairs'', ``Downstairs'' and ``Running'' from USC-HAD dataset. While in HHAR dataset we have samples from activities ``Standing'', ``Sitting'', ``Walking'', ``Upstairs'', ``Downstairs'' and ``Biking''. 
We then down-sample all sensor readings in both datasets to 25Hz, which is a feasible sampling rate on everyday mobile devices with lower overhead. 
After down-sampling, the long consecutive sensor data (up to 5 minutes) are segmented into $6$-second sensor readings. Thus, the HAR task in our experiments is to recognize the activity type for each given $6$-second sample. We have obtained 9,335 and 3,170 6-second samples from the HHAR and USC-HAD datasets, respectively. Finally, we split each of the two datasets into two parts, including training (80\%) and testing (20\%) sets. All models are trained on the entire (or a portion of the) training set and evaluated on the corresponding test set. 
However, in DeepSense on HHAR dataset, apart from sensor readings generated by smart phone, signals from smart watches are also included in their evaluation process. They also augmented the original dataset by adding noise, interpolation and resampling, to get a dataset large enough for training deep neural networks. Therefore, to have a fair comparison to results reported by DeepSense we also evaluate our classification-oriented model on the augmented dataset.  

%\subsection{Experiment setups}
\label{sec:setup}
Experiments with several approaches are conducted to demonstrate the benefits of the proposed similarity embedding network (SEN), including the following algorithms:
\begin{itemize}
  \item \textbf{SEN-SM}: similarity matching on top of the proposed similarity embedding network. 
  \item \textbf{SEN-MLP}: a MLP with one hidden layer, trained separately, applied on top of the proposed similarity embedding network.
  \item \textbf{DeepSense}: the DeepSense model proposed by \citet{yao2017deepsense}.
  \item \textbf{Baseline}: the same network structure as SEN-MLP, yet is trained jointly end-to-end with cross-entropy based on the label prediction error. This is similar to DeepSense in terms of end-to-end cross-entropy training, although differing from DeepSense in detailed network architecture.
\end{itemize}
% introduce setups for stress test
As we have claimed that the proposed SEN can be efficiently trained even with a small number of training samples, we further conduct stress tests to explore the ability of SEN under small training sets. To do this, we randomly choose a small number of samples for each class from the entire raining set, which is 80\% of the original dataset. We then train the SEN and test its HAR performance on the test set with the proposed SEN-SM method. For stress tests, the number of training samples selected varies from 30 to 200.    

Furthermore, to evaluate the robustness of the proposed SEN-SM algorithm to training samples with noisy labels, we perturb the training samples by randomly changing their labels and showing the generalization ability of SEN-SM on a noise-free test set. 
On the USC-HAD dataset, different noise rates, with 10\%--40\% of the samples perturbed, are adopted to evaluate the robustness of the proposed model. 
However, on the HHAR dataset, to avoid showing repeated results, we only evaluate the model when it is trained under the dataset with 40\% of the samples perturbed. 

The accuracy, $F_1$ score and averaged $F_1$ score are widely used in HAR research \cite{bhattacharya2014using,sagha2011benchmarking} for performance evaluation. The averaged $F_1$ score is defined as
\begin{equation} \label{eq:avg_f1}
Avg.F_1 = \frac{\sum_{i = 1}^{c} N_{i} \cdot F_{1}^i }{\sum_{i = 1} ^ {c} N_i},
\end{equation}  
where $c$ is the number of classes, $F_{1} ^ {i}$ is the $F_1$ score of the $i^{th}$ class on the test set and $N_i$ is the number of samples in the $i^{th}$ class. 

In addition, the SEN-SM can be further used for data denoise. We will discuss the denoise techniques and evaluation result in details in Subsection~\ref{exp:denoise}.

\begin{figure}. % show 
  \centering
  \includegraphics[width=2.8in, height = 1.9in ]{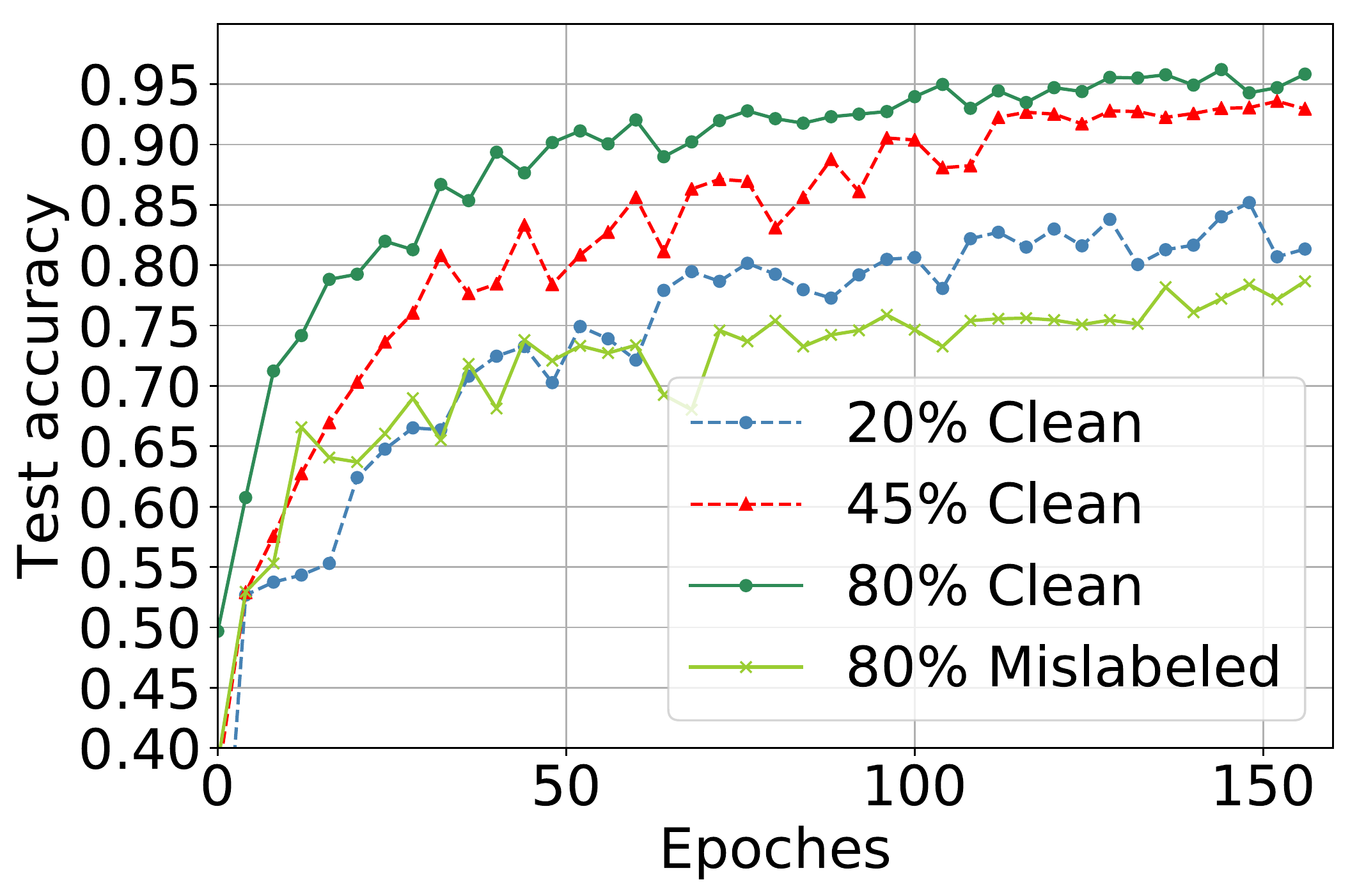}
  \caption{Training curves of Baseline on different sizes of clean and noisy datasets from \emph{HHAR}. 20\%, 45\% and 80\% denote the proportions of the entire HHAR dataset used. ``Clean'' and ``Mislabeled'' indicate whether the training set is clean or noisy. In ``Mislabeled'' dataset, 40\% samples are perturbed.}
  \label{fig:test_acc_noise}
  \Description[training curves on different datasets]{The accuracy curves of models trained on different datasets to show the impact of the number of training samples and corrupted labels.}
\end{figure}

\subsection{HAR Classification results}
To evaluate the efficiency of the proposed network architecture, following the same data processing procedure as DeepSense \cite{deepsense_pro}, we conducted the leave-one-user-out experiment on the augmented HHAR dataset with our Baseline method. Specifically, the DeepSense augment the original dataset ten times by adding Gaussian noise to original signals. On the augmented HHAR dataset, when trained with cross-entropy loss our proposed network, SEN-MLP, can achieve an averaged leave-one-out accuracy over 98\%, which outperforms 94.2\% reported by DeepSense with a great margin. 
 
Then, to show the influence of training dataset on the performance of deep neural network. The Baseline method is evaluated on different sizes of training samples from original HHAR dataset and even training samples with label noise. 
Test accuracy curves are shown in Fig.~\ref{fig:test_acc_noise}, the trend is obvious: as the size of training set grows from 20\% to 80\% the test accuracy improves from around 80\% to near 96\%. Furthermore, 
the model trained on the noise dataset gives the worst performance with an accuracy near 76\% due to the fact that 40\% of its samples are mislabeled. 

\begin{figure}[tbp]
  \centerline{
    \subfigure[Test accuracy on USC-HAD dataset]{
      \includegraphics[width=2in]{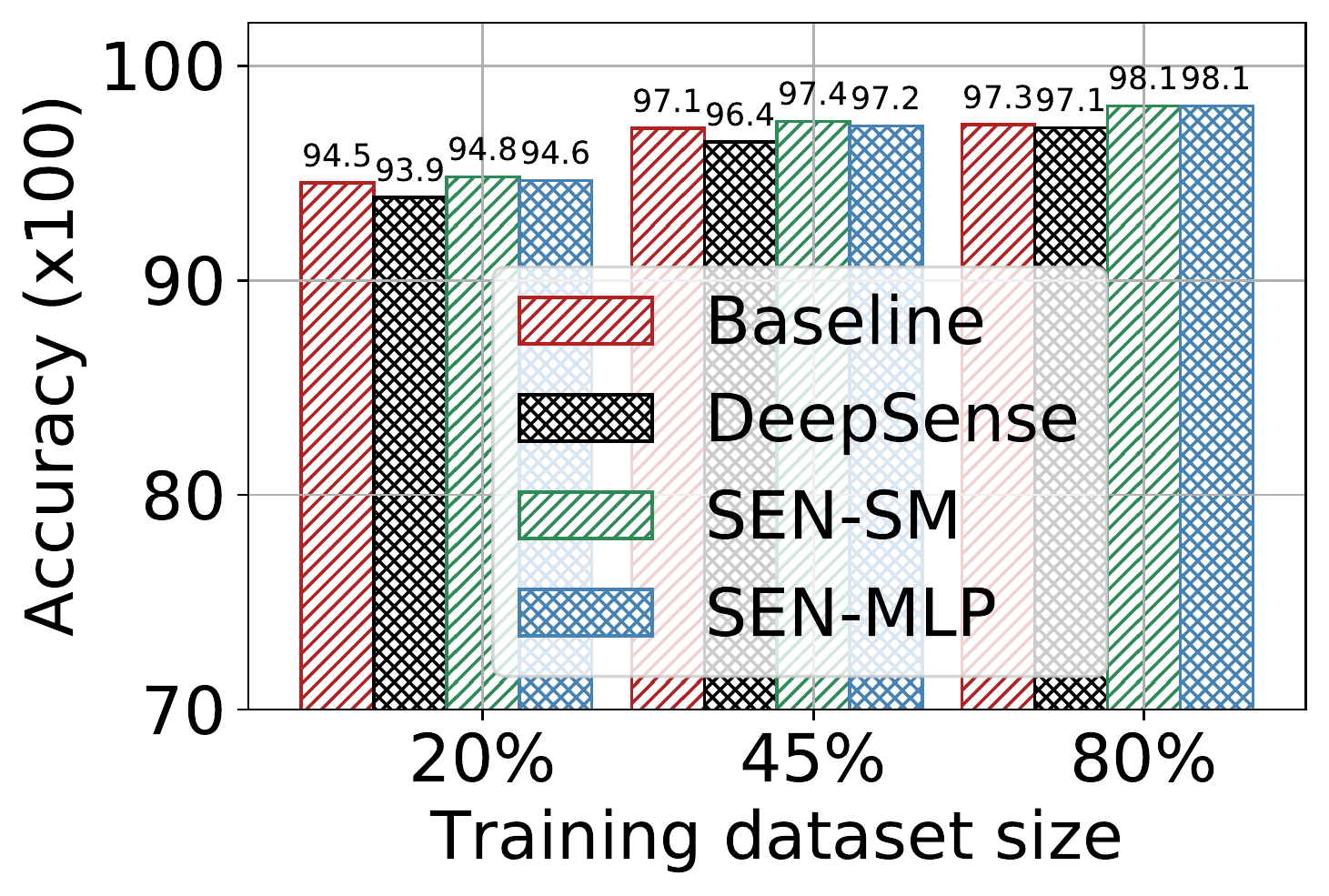}
      \label{fig:acc_three_usc}
    }
    \subfigure[Averaged $F_1$ scores on USC-HAD dataset]{
      \includegraphics[width=2in]{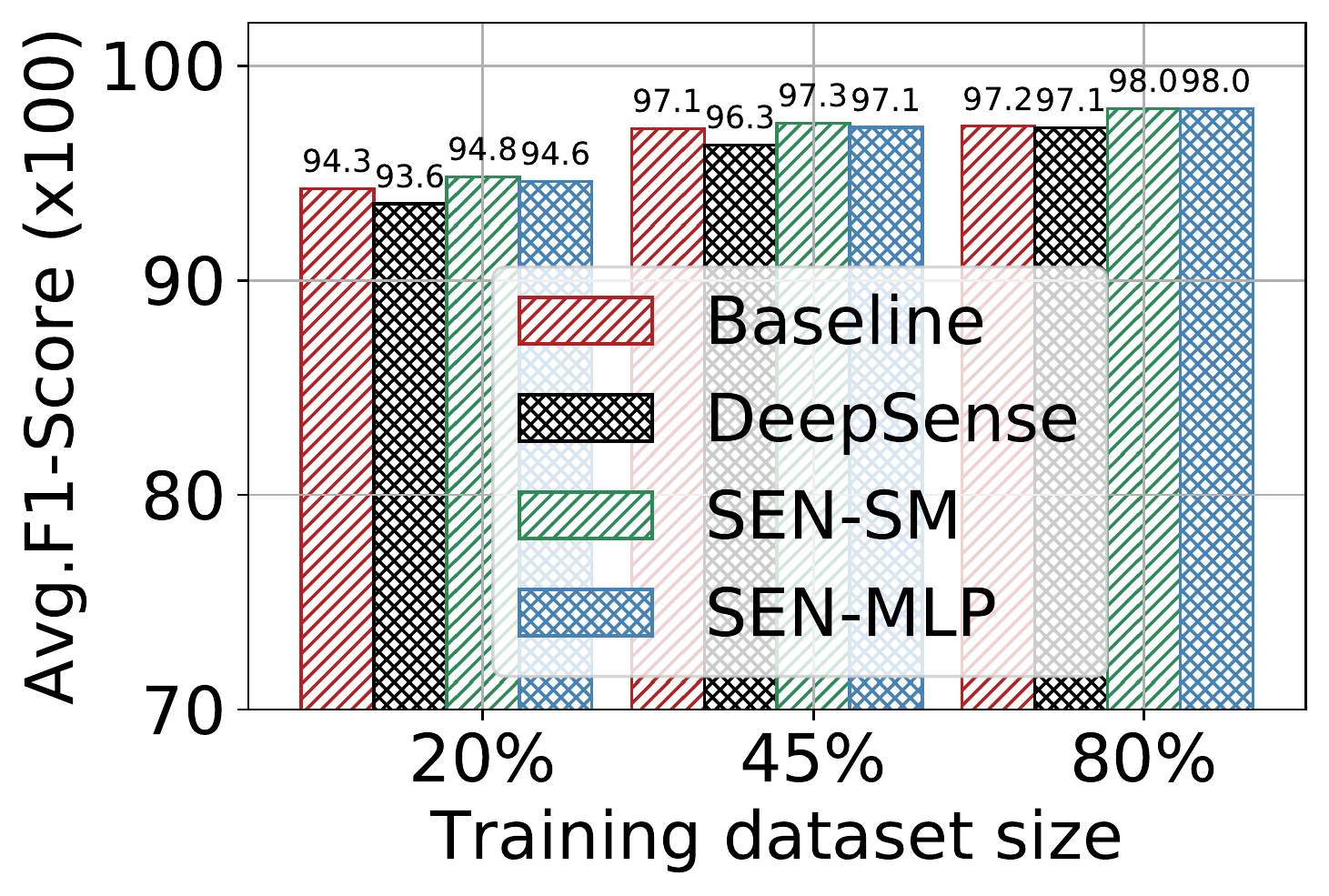}
      \label{fig:f1_three_usc}
    }
  }
  \caption{Test accuracy and averaged $F_1$ scores on USC-HAD dataset of proposed Baseline, SEN-MLP and SEN-SM methods on different sizes of clean training sets. 20\%, 45\% and 80\% represent the proportions of the entire USC-HAD dataset used for training.}
  \label{fig:classification_comp_usc}
  \Description[Test result of proposed method and baselines]{Histogram results on different USC-HAD dataset to show the efficiency of proposed method over baselines.}
\end{figure}

\begin{figure}[tbp]
  \centerline{
    \subfigure[Test accuracy on HHAR dataset]{
      \includegraphics[width=2in]{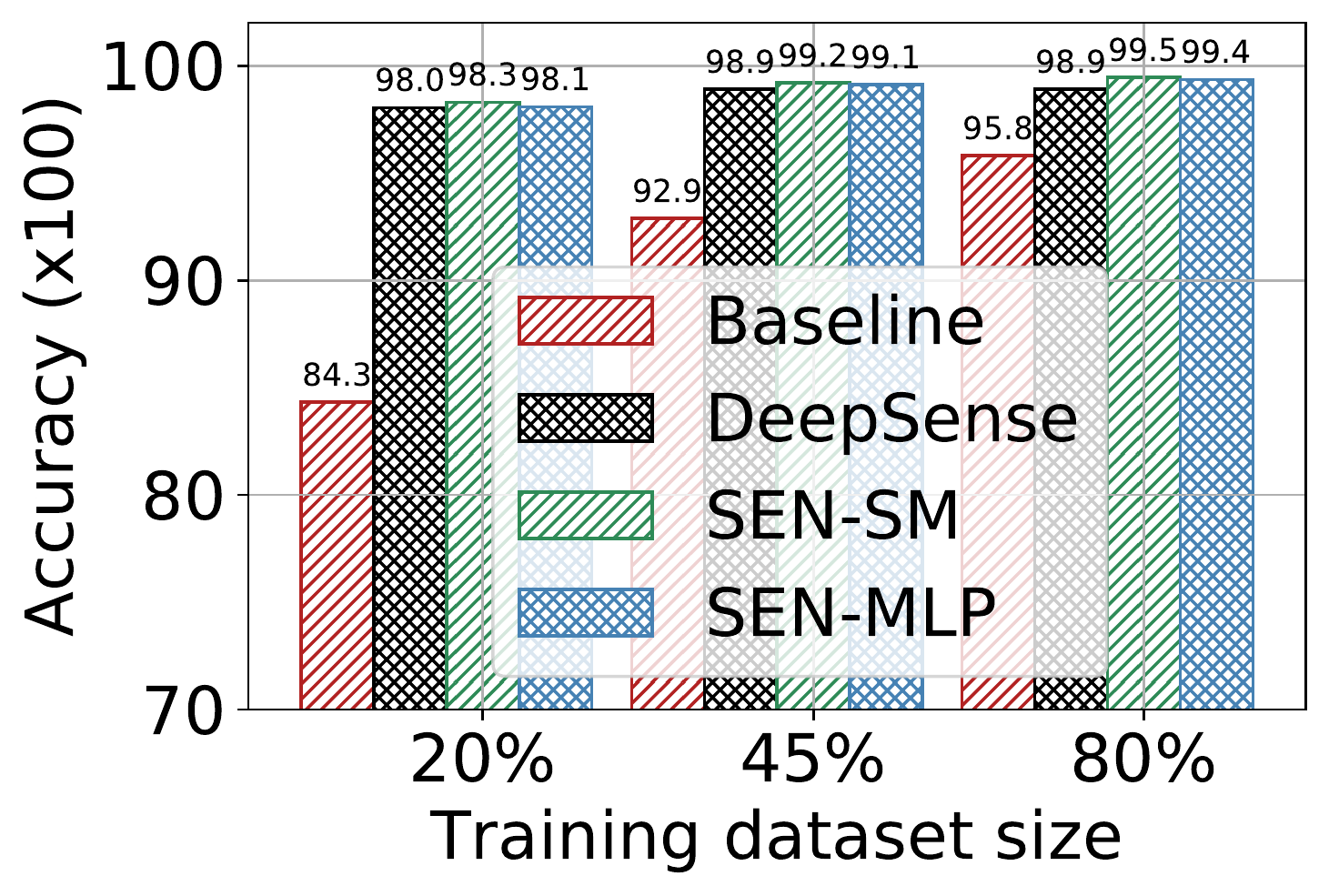}
      \label{fig:acc_three}
    }
    \subfigure[Averaged $F_1$ scores on HHAR dataset]{
      \includegraphics[width=2in]{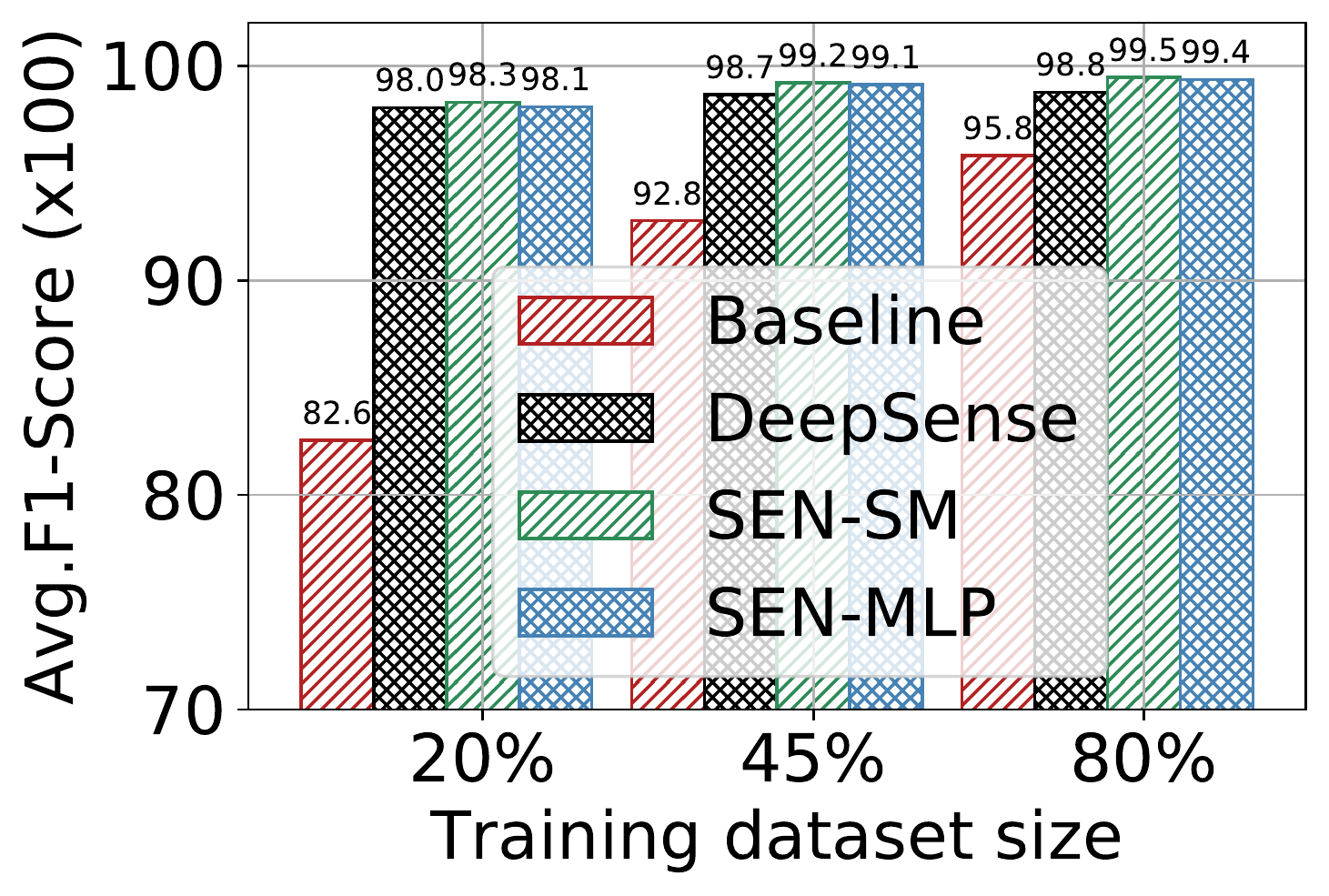}
      \label{fig:f1_three}
    }
  }
  \caption{The accuracy and averaged $F_1$ scores on HHAR dataset of proposed Baseline, SEN-MLP and SEN-SM  methods on different sizes of clean training sets. 20\%, 45\% and 80\% represent the proportions of the entire HHAR dataset used for training.}
  \label{fig:classification_comp}
  \Description[Test result of proposed method and baselines on the HHAR dataset]{Histogram results on different HHAR dataset to show the efficiency of proposed method over baselines.}
\end{figure}

Then we compare the \red{DeepSense and} Baseline algorithm with our proposed SEN-MLP and SEN-SM methods. Detailed results on USC-HAD and HHAR datasets are shown in Fig.~\ref{fig:classification_comp_usc} and Fig.~\ref{fig:classification_comp} respectively. We can see as the training dataset size grows, both the accuracy and  averaged $F_1$ score improve for all the three algorithms. However, SEN-SM and SEN-MLP always achieve superior performance. 
On HHAR dataset, they achieved an accuracy of 98\% and an averaged $F_1$ score above 0.98, even on the smallest dataset (20\%), while the baseline algorithm can only achieve an averaged $F_1$ score of 0.826 and an accuracy of 84.3\%. When the training set size grows to 80\% the SEN-SM and SEN-MLP achieved an accuracy of 99\% with 0.99 averaged $F_1$ score, while the Baseline gives 96\% accuracy and 0.96 averaged $F_1$ score. 
Similar results can also be found on USC-HAD dataset, the SEN-SM and SEN-MLP still outperform Baseline \red{and DeepSense} algorithms with an accuracy above 94\% and an averaged $F_1$ score above 0.94 on 20\% training data.  However, on USC-HAD dataset one can see that the Baseline algorithm also achieved descent performance even on a small training dataset (20\%), with an accuracy of 94.5\% and 0.943 averaged $F_1$ score, this is duo to the fact that the USC-HAD dataset is much simpler than the HHAR dataset regarding to both sensors used and sensing position. 

Note that, on USC-HAD dataset, our Baseline method achieved an accuracy of 97.3\%, comparable to the Deep convolutional neural network (DCNN) \cite{jiang2015human} trained on the original dataset with a much higher sampling frequency (100Hz) with a reported accuracy of 97.1\%, which is significant. \red{Furthermore, on HAR dataset, even through DeepSense is worse than our proposed SEN-SM and SEN-MLP algorithms, it still outperforms our Baseline method which further demonstrates the superior of proposed similarity embedding network.}  

\begin{table}[btp]
\caption{Stress test results on HHAR dataset, the first column shows the number of training samples for each class. }
\begin{center}
\begin{tabular}{lrrr}  
% \toprule
\hline
\# train data& Avg.$F_1$ & Accuracy (\%) & Precision(\%)\\
% \midrule
% \hline
\cline{1-4}
30   (2\%)      & 0.863  & 86.36      & 86.80\\
80   (5\%)      & 0.926  & 92.62      & 92.62\\
110  (7\%)      & 0.950  & 95.03      & 95.14\\
140  (9\%)      & 0.964  & 96.36      & 96.38\\
170  (11\%)     & 0.965  & 96.52      & 96.59\\
200  (13\%)     & 0.978  & 97.81      & 97.82\\
% \bottomrule
\hline
\end{tabular}
\label{tab:stress_all}
\end{center}
\end{table}

\subsection{HAR Stress Testing on Small Training Sets}
To further explore the limit of the proposed SEN, a stress test is conducted on the HHAR dataset, In the stress test, all models are trained on a limited number of training samples for each class and evaluated on the same test dataset. The number of training samples for each class ranges from 30 to 200. 

The averaged $F_1$ score, accuracy and precision scores are shown in Table~\ref{tab:stress_all}. % Fig. ~\ref{fig:wighted_f1_stress} 
As the training dataset grows from 2\% to 13\% the averaged $F_1$ score improved from 0.863 to 0.978. 
One can see, when the number of samples for each class is only $110$ (7\%), we have already achieved an averaged $F_1$ score of 0.95, which is remarkable. 
Note that with only $30 (2\%)$ training samples (each being 6 seconds) available for each class, our method has already achieved an averaged $F_1$ score of 0.86, even better than what we get from the baseline method with 20\% of the training sample, which is 0.826 shown in Fig.~\ref{fig:classification_comp}. This further shows the superior of training embedding networks with pairwise loss compared to cross entropy loss on small training dataset.

\begin{figure}[tbp]
  \centerline{
    \subfigure[Test accuracy]{
      \includegraphics[width=1.6in, height = 1.15in ]{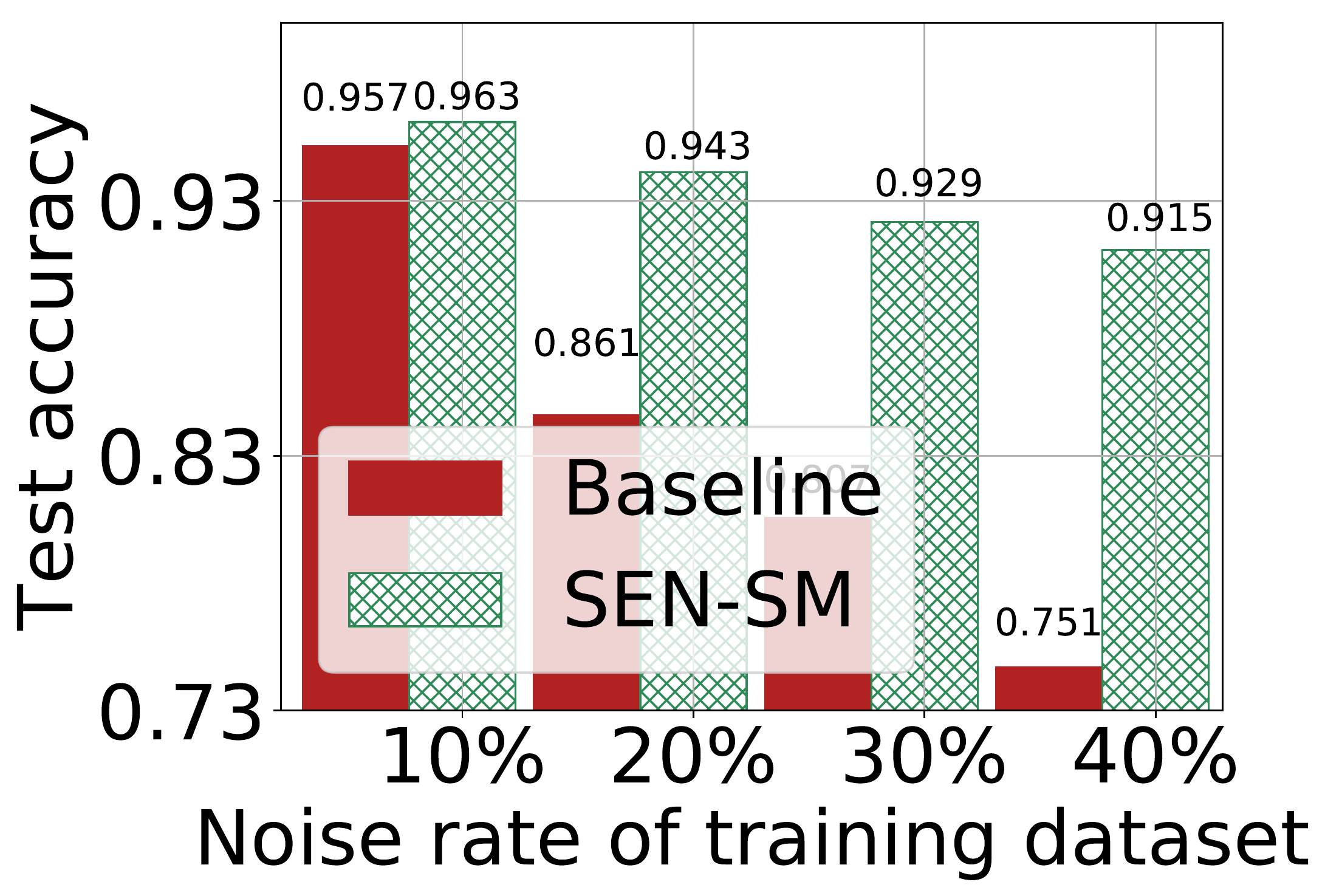}
      \label{fig:acc_three_usc_noise}
    }
    \hspace{3mm}
    \subfigure[Averaged $F_1$ scores]{
      \includegraphics[width=1.6in, height = 1.15in]{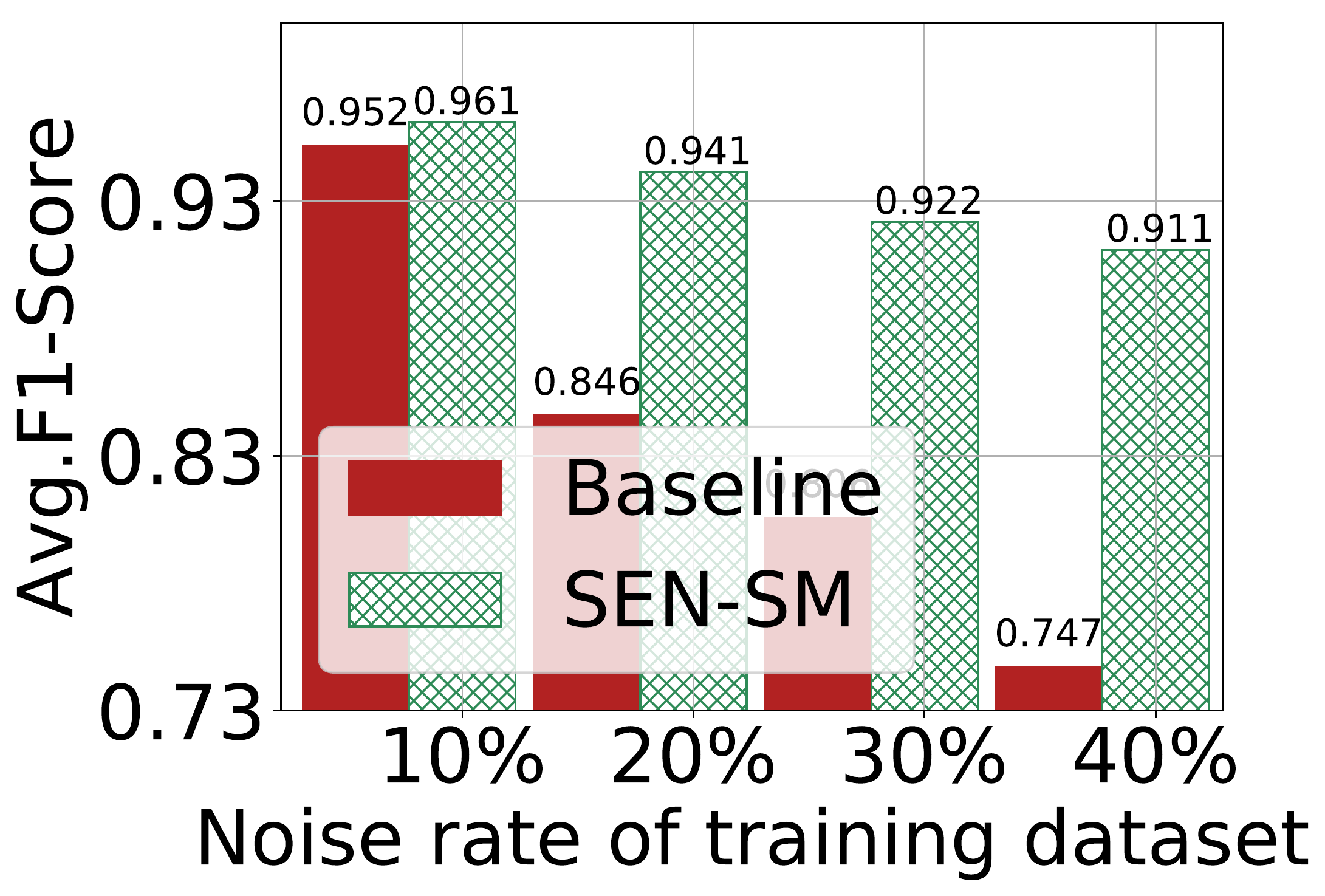}
      \label{fig:f1_three_usc_noise}
    }
  }
  \caption{Test accuracy and averaged $F_1$ scores of proposed SEN-SM method and Baseline algorithm on the same training dataset with different noise rate. 10\%, 20\%, 30\% and 40\% represent the corresponding number of noise samples as their proportions in the whole dataset. Here the training samples are from 80\% of the USC-HAD dataset.}
  \label{fig:noise_robust_usc}
  \Description[Results of all models on noisy dataset]{The superior of proposed method over baselines on noisy dataset with different portion label noise.}
\end{figure}

\begin{figure*}[tbp]
  \centering
    \begin{subfigure}[Stand-Sit]{
      \centering
      \includegraphics[width=1.6in, height = 1.15in]{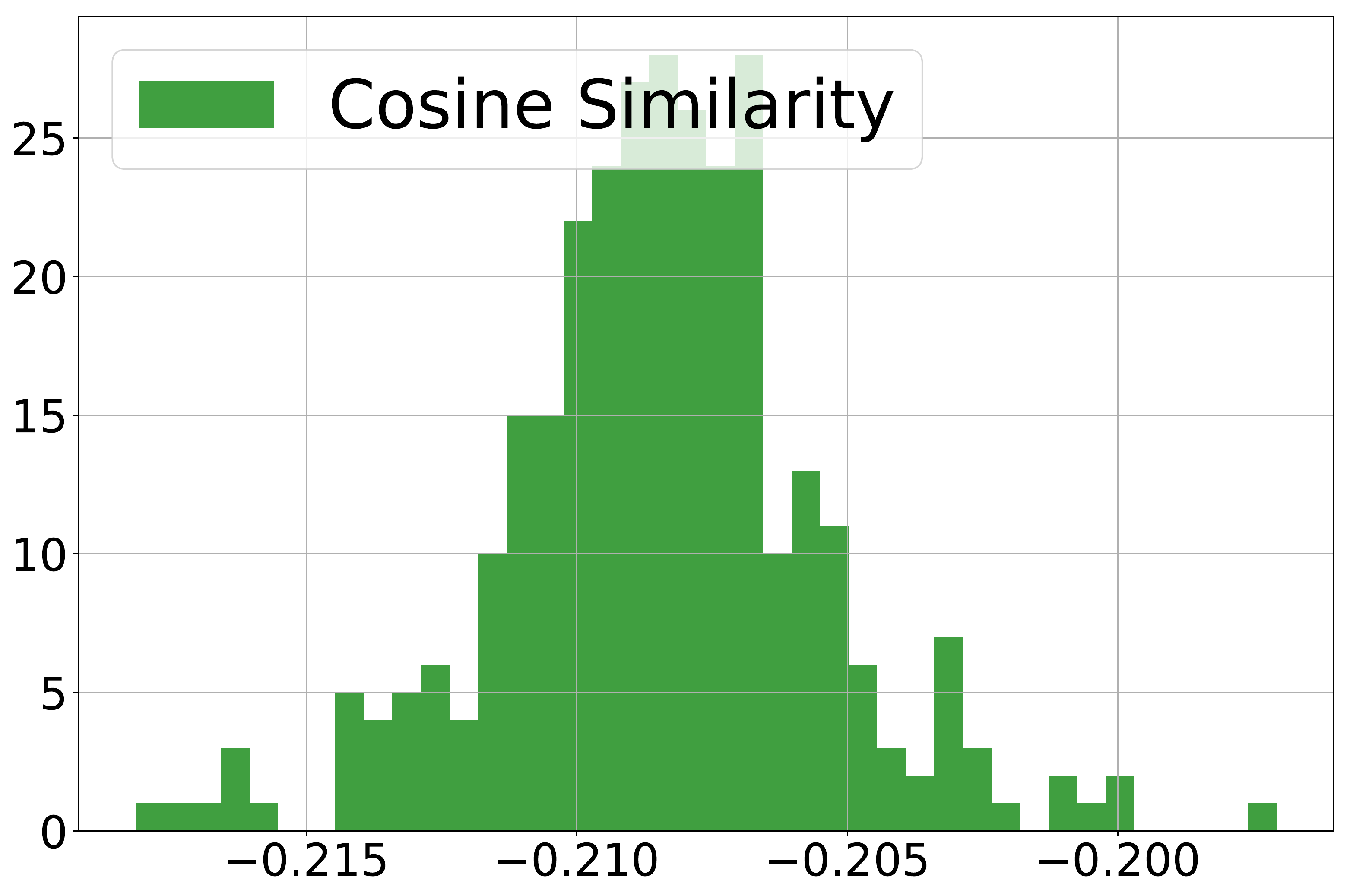}
      \label{fig:stand_sit}
    }
    \end{subfigure}
    \hfill
    \begin{subfigure}[Stand-Walk]{
      \centering
      \includegraphics[width=1.6in, height = 1.15in]{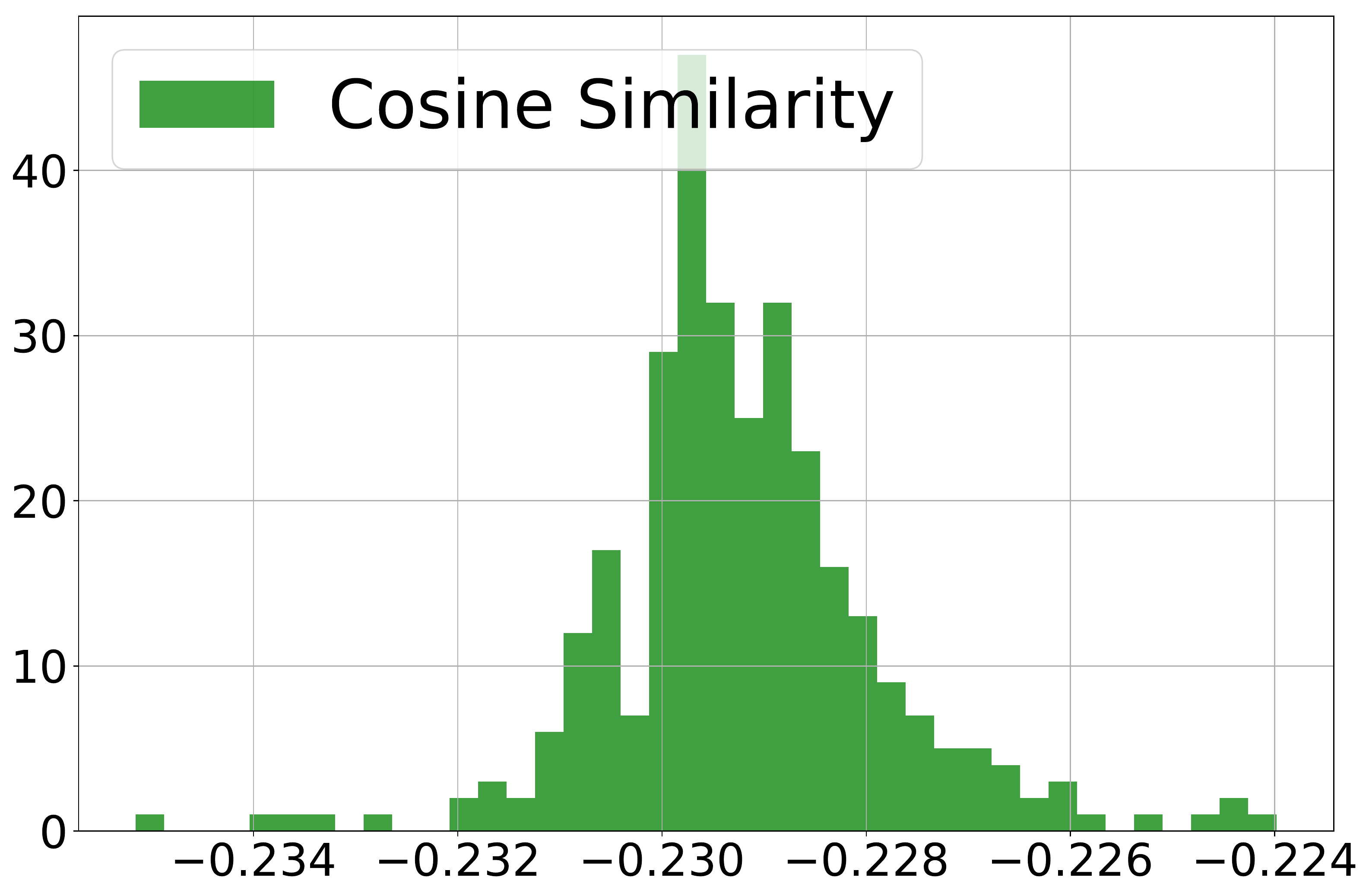}
      \label{fig:stand_walk}
    }
    \end{subfigure}
    % \hspace{3mm}
    % \hfill
    % \begin{subfigure}[Stand-Stairsup]{
    %   \centering
    %   \includegraphics[width=1.6in, height = 1.15in]{fig/Stand_Up.pdf}
    %   \label{fig:stand_up}
    % }
    % \end{subfigure}
    % % \hspace{3mm}
    % \hfill
    % \begin{subfigure}[Stand-Stairsdown]{
    %   \centering
    %   \includegraphics[width=1.6in, height = 1.15in]{fig/Stand_Down.pdf}
    %   \label{fig:stand_down}
    % }
    % \end{subfigure}
    % \hspace{3mm}
    \hfill
    \begin{subfigure}[Stand-Bike]{
      \centering
      \includegraphics[width=1.6in, height = 1.15in]{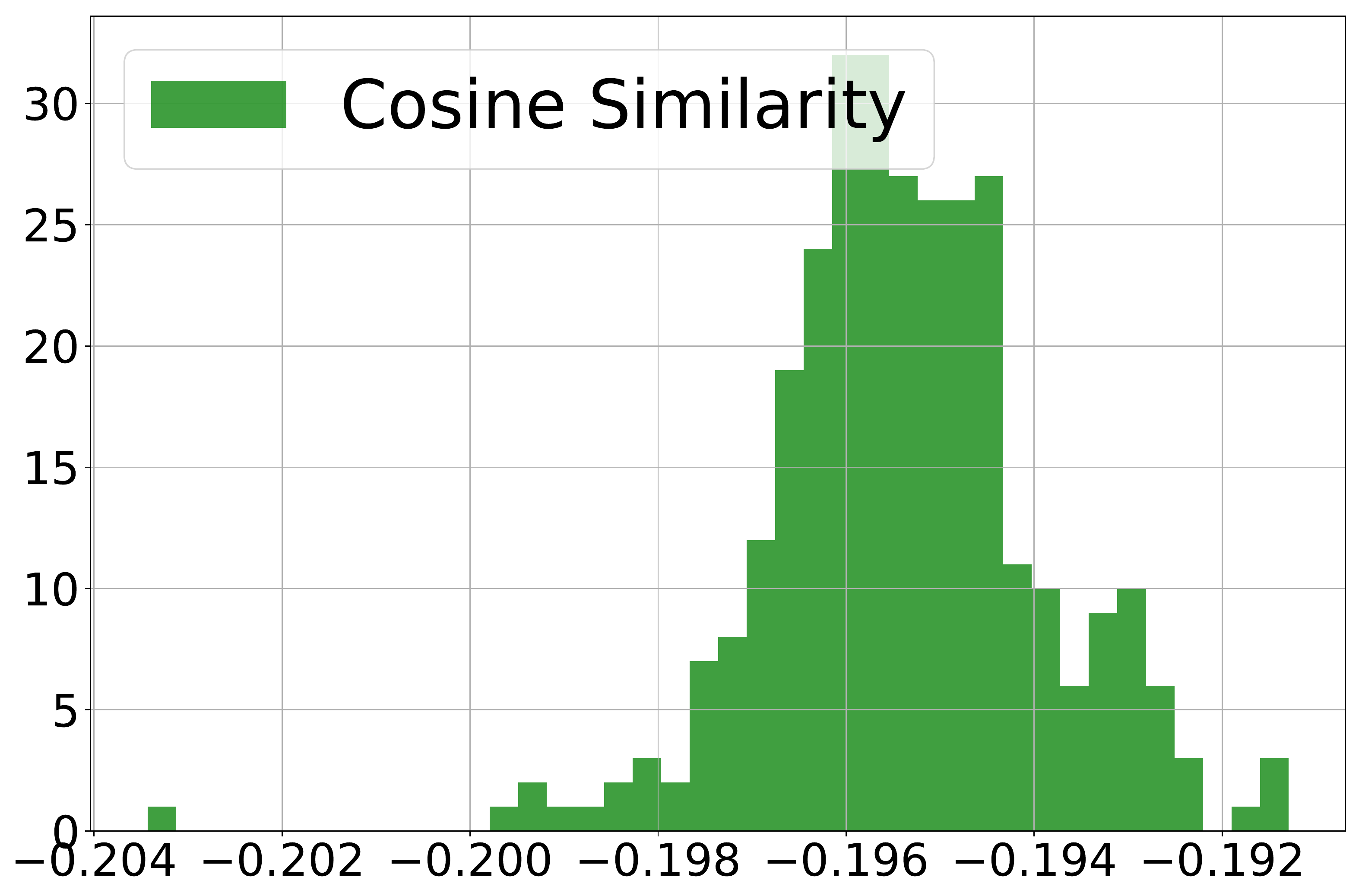}
      \label{fig:stand_bike}
    }
    \end{subfigure}
    % \hspace{3mm}
    \hfill
    \begin{subfigure}[In-Class]{
      \centering
      \includegraphics[width=1.6in, height = 1.15in]{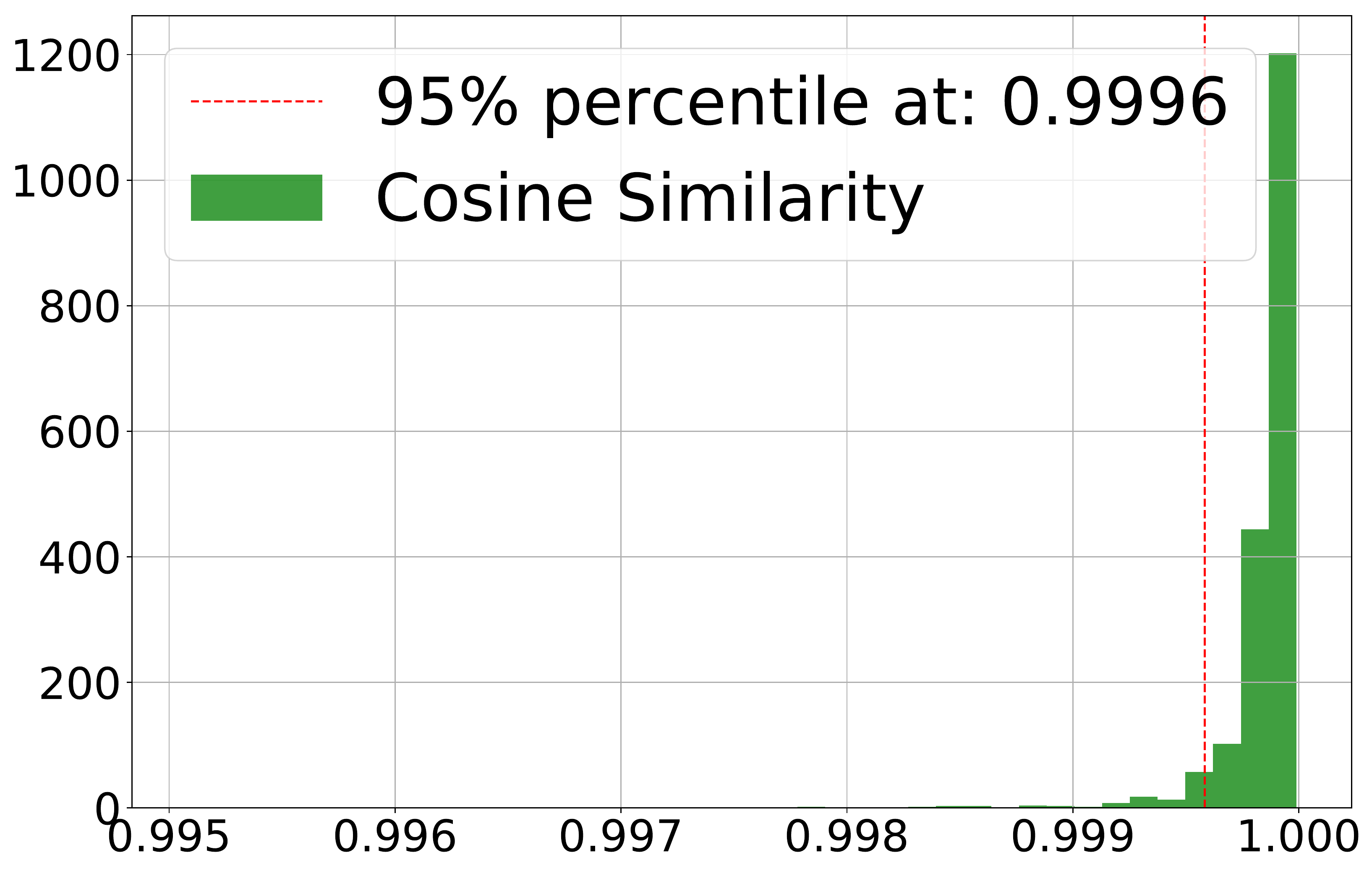}
      \label{fig:inclass}
    }
    \end{subfigure}
    % \hspace{3mm}
    \hfill
    \begin{subfigure}[Cross-class Q-Q Plot]{
      \centering
      \includegraphics[width=1.6in, height = 1.15in]{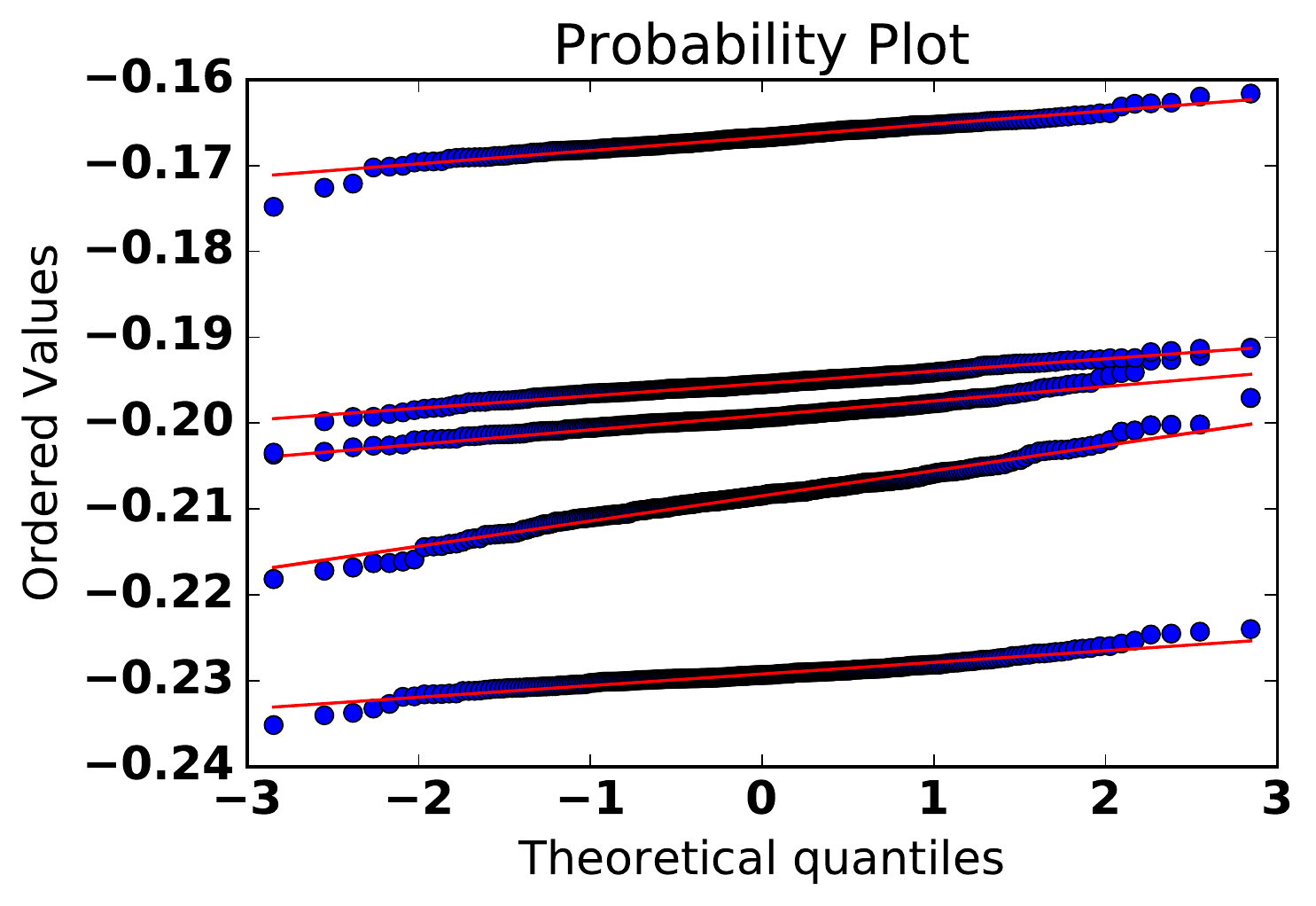}
      \label{fig:between_qq}
    }
    \end{subfigure}
    % \hspace{3mm}
    \hfill
    \begin{subfigure}[In-class Q-Q Plot]{
      \centering
      \includegraphics[width=1.6in, height = 1.3in]{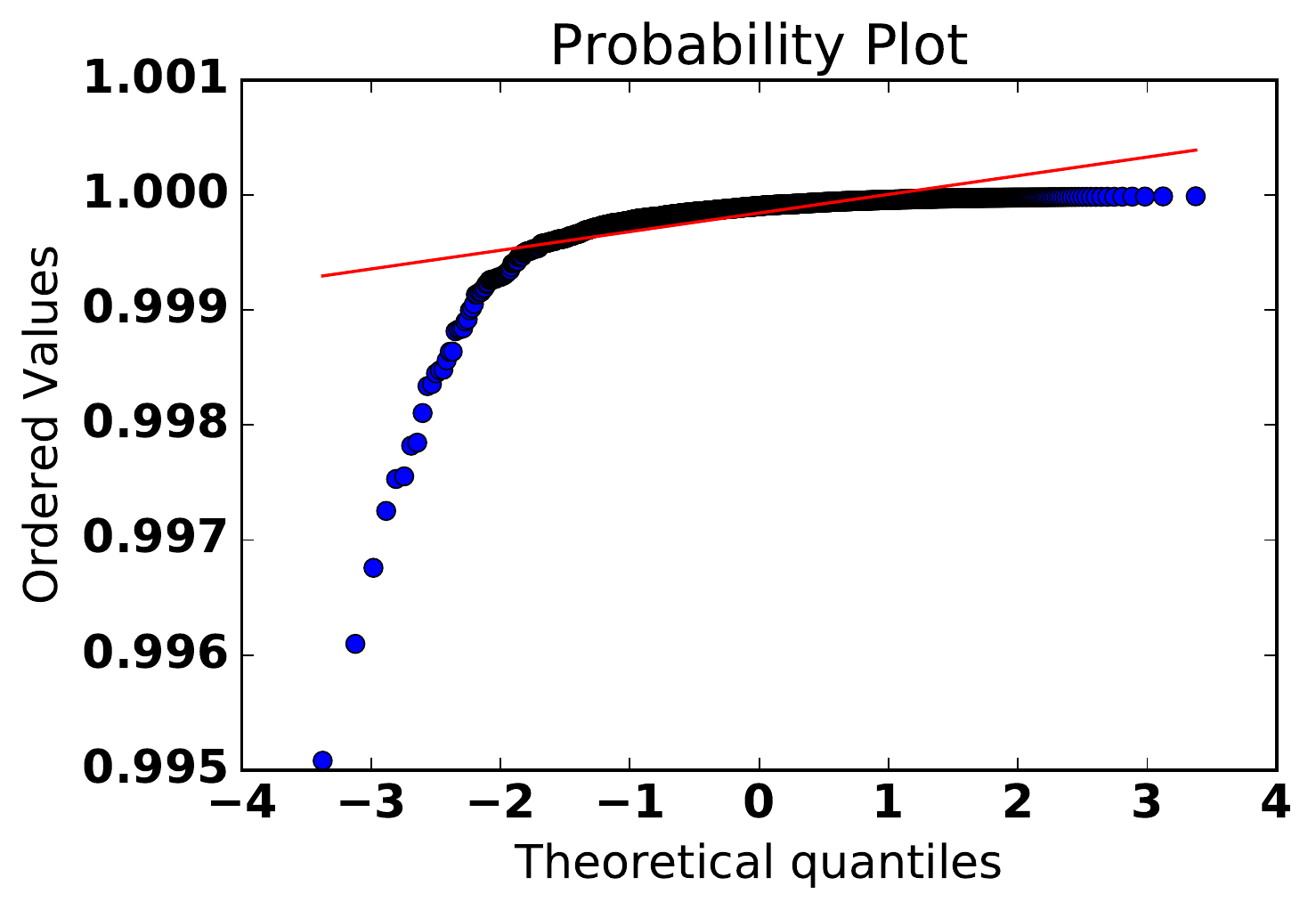}
      \label{fig:inclass_qq}
    }
    \end{subfigure}
  
  \caption{The distribution and Q-Q plot of cosine similarity score. (a)-(c) show the distribution of cosine similarity between samples from ``Stand'' class and class centers of ``Sit'', ``Walk'' and ``Bike''. (d) shows the cosine similarity of all the samples and their corresponding class center. Q-Q plots of between class similarity scores are shown in (e). (f) shows the in-class cosine similarity Q-Q plot.}
  \label{fig:observation}
  \Description[Cosine similarity between different activity types]{Histogram and Q-Q plots are given to show that the Cosine similarity scores between different activity types following a Gaussian distribution.}
\end{figure*}

\subsection{Robustness to Label Noise}
On USC-HAD dataset we trained both the SEN and Baseline model on label noise dataset with different noise rate range from 10\% to 40\% to show the robustness and efficiency of our proposed method in the presence of label noise. The results are shown in Fig.~\ref{fig:noise_robust_usc}. It is obvious that at any noise level, the SEN-SM method achieved better performance that the baseline model. Also, even though the noise rate is only 10\%, the existence of label noise still lead to performance reduction for both SEN-SM and baseline algorithms. As the noise rate grows the performance becoming worse. When the noise rate grows from 10\% to 40\%, the test accuracy of the baseline method the decline is much dramatic from 95.7\% to 75.1\% while the SEN-SM model reduced from 96.3\% to 91.5\%.  

On HHAR dataset, we evaluated both algorithms on a noisy dataset with 40\% noise rate, the SEN-SM achieves an averaged $F_1$ score of 0.84 with a 84.48\% accuracy, while the baseline giving an averaged $F_1$ score of 0.73 and 77.25\% accuracy. Compared to the original 99\% and 96\% accuracy achieved on noise free dataset, shown in Fig.~\ref{fig:classification_comp}, our proposed model got less accuracy reduction and much better overall performance which shows the robustness of our proposed SEN-SM method. 

\subsection{Data Denoising}
\label{exp:denoise}
Our proposed similarity embedding network can also be used to denoise a large noisy dataset, when trained on a small clean dataset. The goal of data denoising is to remove as many wrongly labeled samples as possible, so that the denoised data can be leveraged for training HAR models or other tasks. 

We found that the between-class cosine distances (cosine similarities), i.e., cosine distances between each sample embedding and other class centers, follow a Gaussian distribution while the in-class cosine distances, i.e., the cosine distance between each sample embedding and its own class center, highly concentrate around 1 as is suggested by Fig.~\ref{fig:stand_sit}-Fig.~\ref{fig:inclass}. The $Q-Q$ plots are provided to further show their distributions in Fig.~\ref{fig:between_qq} and Fig.~\ref{fig:inclass_qq}. 

With this observation, we propose to filter out possibly mislabeled samples as follows. For any given sample, we consider it as correctly labeled if and only if 
\begin{itemize}
  \item its embedding is close enough to its own class center, i.e., its in-class cosine distance is no smaller than 5\% of all the in-class cosine distances in that class;
  \item its embedding is far from all the other class centers, i.e., 
its cosine distance to another class center is smaller than $\mu + 2\sigma$, where $\mu$ and $\sigma$ are the mean and standard deviation of the corresponding between-class cosine distances. 
\end{itemize} 
In this manner, we can filter out noise samples with a low miss rate (high recall). Although some clean samples may also be filtered out, the idea of denoising is to detect as many mislabeled samples as possible to make sure the retrieved data is clean. 

\begin{table}[btp]
\caption{Denoise results on HHAR dataset, the first column shows the number of training samples for each class.}
\begin{center}
\begin{tabular}{lrrr}  
% \toprule
\hline
\# train data& Avg.$F_1$ & Accuracy (\%) & Recall(x100)\\
% \midrule
% \hline
\cline{1-4}
30   (2\%)      & 0.90   & 88.01      & 99.05\\
80   (5\%)      & 0.937  & 92.78      & 99.51\\
110  (7\%)      & 0.957  & 95.17      & 99.80\\
140  (9\%)      & 0.954  & 94.77      & 99.71\\
170  (11\%)     & 0.972  & 96.86      & 99.80\\
200  (13\%)     & 0.978  & 97.61      & 99.77\\
% \bottomrule
\hline
\end{tabular}
\label{tab:denoise}
\end{center}
\end{table}

To evaluate denoising performance, we first train our SEN model on a randomly selected small dataset. Then, the in-class and between-class cosine similarity distance distributions are approximated on the same dataset. Finally, with the trained embedding neural network, we calculated class centers as well as the estimated statistics we can denoise a large, noisy dataset with high recall score. 

We evaluate this property by trying different number of clean samples selected from HHAR dataset. Similar to stress tests mentioned above, we vary the number of samples for each class from 30 to 200 in the training set.
Apart from the metrics mentioned above, recall score, which shows the detection rate of positive samples, is an important criterion when it comes to data denoise problem, thus is adopted for the evaluation of data denoise performance.

We evaluate the denoise performance with the manually perturbed labels as noise on HHAR dataset. As shown in Table ~\ref{tab:denoise}, the evaluation is also done on different sizes of clean training datasets like what we did in stress test. As one can see, the larger the training set is the better the denoise performance. 
All the recall scores are close to 1 (almost no miss) while the denoise accuracy and $F_1$ score grow as the training dataset size grows from 30 to 200 samples per class. The $F_1$ scores and accuracy improved from 0.9 and 87\% to 0.972 and 97\% when the training dataset size grows from 30 to 200 samples per class, which also means that fewer clean samples will be mistakenly detected as noise. 

% related work
\section{Related work}
\label{sec:related}
Some recent studies also apply deep neural network models to mobile sensing or HAR applications. 
\cite{bhattacharya2016smart} and \cite{radu2016towards} use deep Boltzmann Machine and Multimodal DBMs to improve the performance of heterogeneous human activity recognition. 
DeepEar \cite{lane2015deepear} also uses Deep Boltzmann Machine to improve the performance of audio sensing tasks in an environment with background noise. 
IDNet \cite{gadaleta2018idnet} applies CNNs to the biometric gait analysis task. 
DeepX \cite{lane2016deepx} and RedEye \cite{likamwa2016redeye} reduce the energy consumption of deep neural networks, based on software and hardware, respectively. However, these studies do not capture the temporal relationship in time-series sensor inputs, and, with the only exception of the MultiRBM, lack the capability of fusing multimodal sensor input. 
\citet{zhu2014human} calculates the similarity degree between handcrafted recognition feature and activity features to identify the most possible phone position, then the result of this similarity matching is used for further activity recognition.  
Deepsense \cite{yao2017deepsense} applied RNN on top of CNN to acquire the sequential information of the input sensor data. \red{\citet{wan2020deep} designs a smartphone inertial accelerometer-based architecture for HAR which also use CNN and RNN modules for better prediction accuracy. }
Those works all give out a classification model for HAR or other context awareness applications and can not be used  for denoise and have limited generalization performance when training data is insufficient or even noisy with wrong labels. Recently, 
\red{\citet{chen2019semisupervised} use a semi-supervised approach to deal with class imbalance in small labeled datasets. \citet{bai2019motion2vector} proposed a unsupervised learning method, Motion2Vector, to convert a time interval of activity sensor data into a movement vector embedding in a multidimensional space. 
Instead of using unsupervised or semi-supervised training scheme, we adopt pairwise loss to train our proposed similarity embedding network which can also alleviate the problem caused by a small amount of training samples.}

Learning to hash has been attracting a large amount of research interest in machine learning and computer vision \cite{wang2018survey}. 
Applications of learning to hash including large scale object retrieval \cite{jegou2010aggregating}, image classification \cite{sanchez2011high} and detection \cite{vedaldi2012sparse}. 
Similarity preserving is the main methodology of learning to hash. 
Recent deep learning based studies in learning to hash use deep neural networks to simultaneously learn the image representation and approximate the hashing function.
Convolutional Neural Network Hashing (CNNH) [20] is one of the early works to incorporate deep neural networks into hash coding
Supervised Discrete Hashing (SDH) \cite{shen2015supervised}, which directly optimize the binary hash codes via the discrete cyclic coordinate descend method. 
The most relevant work is Deep Supervised Discrete Hashing (DSDH) \cite{li2017deep} which uses CNN to learn the image representation and hash function simultaneously and output the binary hashing encoding directly. However, DSDH uses hamming distance based pairwise loss and prediction loss as objective while we directly minimize the cosine distance based pairwise loss. 

Some studies in metric learning that try to learn a transformation from input space to a low-dimension feature space \cite{atkeson1997locally} are also similar to our work. 
Neighborhood Component Analysis (NCA) \cite{goldberger2005neighbourhood} learn a low-dimensional linear embedding of labeled data by directly maximizing a stochastic variant of the leave-one-out KNN. 
\cite{salakhutdinov2007learning} proposed methods to pretrain and fine-tune a multilayer neural network to learn a nonlinear transformation from input to feature space where nonparametric classification methods performs well. However, both the pretrain and fine-tune are done in an unsupervised manner while in our network the training of the embedding network is supervised with pairwise similarity.
\cite{vinyals2016matching} adopts a support set $S$ for one-shot learning but is trained with the prediction loss of labels. 

%  Conclusion
\section{Conclusion}
\label{sec:conclude}
In this paper, we propose a robust similarity embedding network to solve the main challenges faced when deploying complex deep models for HAR tasks in real world. 
Our similarity embedding networks, even trained with cross entropy loss on a dataset with much lower sampling rate, can already achieve a performance comparable to or even outperform most of the state-of-the-art algorithms for HAR problem.
By adopting pairwise loss, our model can generalize well even if trained on a small dataset or noisy data with mislabeled samples. Extensive experimental results on two publicly available datasets have demonstrated the superiority of our proposed embedding network model. Stress tests on a heterogeneous dataset shows the robustness of SEN to small training sets, while experiments on noisy datasets have shown its robustness to labeling noise in the training data. These capabilities are missing in the existing deep neural network models. Finally, even trained on a small clean dataset, the proposed SEN is capable of denoising a heavily contaminated larger dataset with a 40\% noise rate.

\bibliographystyle{ACM-Reference-Format}
\bibliography{main}


\begin{thebibliography}{44}

%%% the Web address.

\ifx \showCODEN    \undefined \def \showCODEN     #1{\unskip}     \fi
\ifx \showDOI      \undefined \def \showDOI       #1{#1}\fi
\ifx \showISBNx    \undefined \def \showISBNx     #1{\unskip}     \fi
\ifx \showISBNxiii \undefined \def \showISBNxiii  #1{\unskip}     \fi
\ifx \showISSN     \undefined \def \showISSN      #1{\unskip}     \fi
\ifx \showLCCN     \undefined \def \showLCCN      #1{\unskip}     \fi
\ifx \shownote     \undefined \def \shownote      #1{#1}          \fi
\ifx \showarticletitle \undefined \def \showarticletitle #1{#1}   \fi
\ifx \showURL      \undefined \def \showURL       {\relax}        \fi
% invisible to TeX
\providecommand\bibfield[2]{#2}
\providecommand\bibinfo[2]{#2}
\providecommand\natexlab[1]{#1}
\providecommand\showeprint[2][]{arXiv:#2}

\bibitem[\protect\citeauthoryear{Alemdar and Ersoy}{Alemdar and Ersoy}{2010}]%
        {alemdar2010wireless}
\bibfield{author}{\bibinfo{person}{Hande Alemdar} {and} \bibinfo{person}{Cem
  Ersoy}.} \bibinfo{year}{2010}\natexlab{}.
\newblock \showarticletitle{Wireless sensor networks for healthcare: A survey}.
\newblock \bibinfo{journal}{\emph{Computer networks}} \bibinfo{volume}{54},
  \bibinfo{number}{15} (\bibinfo{year}{2010}), \bibinfo{pages}{2688--2710}.
\newblock


\bibitem[\protect\citeauthoryear{Anguita, Ghio, Oneto, Parra, and
  Reyes-Ortiz}{Anguita et~al\mbox{.}}{2012}]%
        {svm12har}
\bibfield{author}{\bibinfo{person}{Davide Anguita}, \bibinfo{person}{Alessandro
  Ghio}, \bibinfo{person}{Luca Oneto}, \bibinfo{person}{Xavier Parra}, {and}
  \bibinfo{person}{Jorge~L. Reyes-Ortiz}.} \bibinfo{year}{2012}\natexlab{}.
\newblock \showarticletitle{Human Activity Recognition on Smartphones Using a
  Multiclass Hardware-Friendly Support Vector Machine}. In
  \bibinfo{booktitle}{\emph{Ambient Assisted Living and Home Care}},
  \bibfield{editor}{\bibinfo{person}{Jos{\'e} Bravo},
  \bibinfo{person}{Ram{\'o}n Herv{\'a}s}, {and} \bibinfo{person}{Marcela
  Rodr{\'i}guez}} (Eds.). \bibinfo{publisher}{Springer Berlin Heidelberg},
  \bibinfo{pages}{216--223}.
\newblock
\showISBNx{978-3-642-35395-6}


\bibitem[\protect\citeauthoryear{Atkeson, Moore, and Schaal}{Atkeson
  et~al\mbox{.}}{1997}]%
        {atkeson1997locally}
\bibfield{author}{\bibinfo{person}{Christopher~G Atkeson},
  \bibinfo{person}{Andrew~W Moore}, {and} \bibinfo{person}{Stefan Schaal}.}
  \bibinfo{year}{1997}\natexlab{}.
\newblock \showarticletitle{Locally weighted learning for control}.
\newblock In \bibinfo{booktitle}{\emph{Lazy learning}}.
  \bibinfo{publisher}{Springer}, \bibinfo{pages}{75--113}.
\newblock


\bibitem[\protect\citeauthoryear{Bai, Yeung, Efstratiou, and Chikomo}{Bai
  et~al\mbox{.}}{2019}]%
        {bai2019motion2vector}
\bibfield{author}{\bibinfo{person}{Lu Bai}, \bibinfo{person}{Chris Yeung},
  \bibinfo{person}{Christos Efstratiou}, {and} \bibinfo{person}{Moyra
  Chikomo}.} \bibinfo{year}{2019}\natexlab{}.
\newblock \showarticletitle{Motion2Vector: unsupervised learning in human
  activity recognition using wrist-sensing data}. In
  \bibinfo{booktitle}{\emph{Adjunct Proceedings of the 2019 ACM International
  Joint Conference on Pervasive and Ubiquitous Computing and Proceedings of the
  2019 ACM International Symposium on Wearable Computers}}.
  \bibinfo{pages}{537--542}.
\newblock


\bibitem[\protect\citeauthoryear{Bhattacharya and Lane}{Bhattacharya and
  Lane}{2016}]%
        {bhattacharya2016smart}
\bibfield{author}{\bibinfo{person}{Sourav Bhattacharya} {and}
  \bibinfo{person}{Nicholas~D Lane}.} \bibinfo{year}{2016}\natexlab{}.
\newblock \showarticletitle{From smart to deep: Robust activity recognition on
  smartwatches using deep learning}. In \bibinfo{booktitle}{\emph{Pervasive
  Computing and Communication Workshops (PerCom Workshops), 2016 IEEE
  International Conference on}}. IEEE, \bibinfo{pages}{1--6}.
\newblock


\bibitem[\protect\citeauthoryear{Bhattacharya, Nurmi, Hammerla, and
  Pl{\"o}tz}{Bhattacharya et~al\mbox{.}}{2014}]%
        {bhattacharya2014using}
\bibfield{author}{\bibinfo{person}{Sourav Bhattacharya},
  \bibinfo{person}{Petteri Nurmi}, \bibinfo{person}{Nils Hammerla}, {and}
  \bibinfo{person}{Thomas Pl{\"o}tz}.} \bibinfo{year}{2014}\natexlab{}.
\newblock \showarticletitle{Using unlabeled data in a sparse-coding framework
  for human activity recognition}.
\newblock \bibinfo{journal}{\emph{Pervasive and Mobile Computing}}
  \bibinfo{volume}{15} (\bibinfo{year}{2014}), \bibinfo{pages}{242--262}.
\newblock


\bibitem[\protect\citeauthoryear{Bulling, Blanke, and Schiele}{Bulling
  et~al\mbox{.}}{2014}]%
        {bulling2014tutorial}
\bibfield{author}{\bibinfo{person}{Andreas Bulling}, \bibinfo{person}{Ulf
  Blanke}, {and} \bibinfo{person}{Bernt Schiele}.}
  \bibinfo{year}{2014}\natexlab{}.
\newblock \showarticletitle{A tutorial on human activity recognition using
  body-worn inertial sensors}.
\newblock \bibinfo{journal}{\emph{ACM Computing Surveys (CSUR)}}
  \bibinfo{volume}{46}, \bibinfo{number}{3} (\bibinfo{year}{2014}),
  \bibinfo{pages}{33}.
\newblock


\bibitem[\protect\citeauthoryear{Chen, Yao, Zhang, Wang, Chang, and Nie}{Chen
  et~al\mbox{.}}{2019}]%
        {chen2019semisupervised}
\bibfield{author}{\bibinfo{person}{Kaixuan Chen}, \bibinfo{person}{Lina Yao},
  \bibinfo{person}{Dalin Zhang}, \bibinfo{person}{Xianzhi Wang},
  \bibinfo{person}{Xiaojun Chang}, {and} \bibinfo{person}{Feiping Nie}.}
  \bibinfo{year}{2019}\natexlab{}.
\newblock \showarticletitle{A semisupervised recurrent convolutional attention
  model for human activity recognition}.
\newblock \bibinfo{journal}{\emph{IEEE transactions on neural networks and
  learning systems}} \bibinfo{volume}{31}, \bibinfo{number}{5}
  (\bibinfo{year}{2019}), \bibinfo{pages}{1747--1756}.
\newblock


\bibitem[\protect\citeauthoryear{Gadaleta and Rossi}{Gadaleta and
  Rossi}{2018}]%
        {gadaleta2018idnet}
\bibfield{author}{\bibinfo{person}{Matteo Gadaleta} {and}
  \bibinfo{person}{Michele Rossi}.} \bibinfo{year}{2018}\natexlab{}.
\newblock \showarticletitle{Idnet: Smartphone-based gait recognition with
  convolutional neural networks}.
\newblock \bibinfo{journal}{\emph{Pattern Recognition}}  \bibinfo{volume}{74}
  (\bibinfo{year}{2018}), \bibinfo{pages}{25--37}.
\newblock


\bibitem[\protect\citeauthoryear{Goldberger, Hinton, Roweis, and
  Salakhutdinov}{Goldberger et~al\mbox{.}}{2005}]%
        {goldberger2005neighbourhood}
\bibfield{author}{\bibinfo{person}{Jacob Goldberger},
  \bibinfo{person}{Geoffrey~E Hinton}, \bibinfo{person}{Sam~T Roweis}, {and}
  \bibinfo{person}{Ruslan~R Salakhutdinov}.} \bibinfo{year}{2005}\natexlab{}.
\newblock \showarticletitle{Neighbourhood components analysis}. In
  \bibinfo{booktitle}{\emph{Advances in neural information processing
  systems}}. \bibinfo{pages}{513--520}.
\newblock


\bibitem[\protect\citeauthoryear{Hammerla, Halloran, and Pl{\"o}tz}{Hammerla
  et~al\mbox{.}}{2016}]%
        {hammerla2016deep}
\bibfield{author}{\bibinfo{person}{Nils~Y Hammerla}, \bibinfo{person}{Shane
  Halloran}, {and} \bibinfo{person}{Thomas Pl{\"o}tz}.}
  \bibinfo{year}{2016}\natexlab{}.
\newblock \showarticletitle{Deep, Convolutional, and Recurrent Models for Human
  Activity Recognition Using Wearables}. In \bibinfo{booktitle}{\emph{IJCAI}}.
\newblock


\bibitem[\protect\citeauthoryear{Hu, Liu, Su, Wang, Abdelzaher, Hui, Zheng,
  Xie, and Stankovic}{Hu et~al\mbox{.}}{2014}]%
        {hu2014towards}
\bibfield{author}{\bibinfo{person}{Shaohan Hu}, \bibinfo{person}{Hengchang
  Liu}, \bibinfo{person}{Lu Su}, \bibinfo{person}{Hongyan Wang},
  \bibinfo{person}{Tarek~F Abdelzaher}, \bibinfo{person}{Pan Hui},
  \bibinfo{person}{Wei Zheng}, \bibinfo{person}{Zhiheng Xie}, {and}
  \bibinfo{person}{John~A Stankovic}.} \bibinfo{year}{2014}\natexlab{}.
\newblock \showarticletitle{Towards automatic phone-to-phone communication for
  vehicular networking applications}. In \bibinfo{booktitle}{\emph{INFOCOM,
  2014 Proceedings IEEE}}. IEEE, \bibinfo{pages}{1752--1760}.
\newblock


\bibitem[\protect\citeauthoryear{J{\'e}gou, Douze, Schmid, and
  P{\'e}rez}{J{\'e}gou et~al\mbox{.}}{2010}]%
        {jegou2010aggregating}
\bibfield{author}{\bibinfo{person}{Herv{\'e} J{\'e}gou},
  \bibinfo{person}{Matthijs Douze}, \bibinfo{person}{Cordelia Schmid}, {and}
  \bibinfo{person}{Patrick P{\'e}rez}.} \bibinfo{year}{2010}\natexlab{}.
\newblock \showarticletitle{Aggregating local descriptors into a compact image
  representation}. In \bibinfo{booktitle}{\emph{Computer Vision and Pattern
  Recognition (CVPR), 2010 IEEE Conference on}}. IEEE,
  \bibinfo{pages}{3304--3311}.
\newblock


\bibitem[\protect\citeauthoryear{Jiang and Yin}{Jiang and Yin}{2015}]%
        {jiang2015human}
\bibfield{author}{\bibinfo{person}{Wenchao Jiang} {and}
  \bibinfo{person}{Zhaozheng Yin}.} \bibinfo{year}{2015}\natexlab{}.
\newblock \showarticletitle{Human activity recognition using wearable sensors
  by deep convolutional neural networks}. In
  \bibinfo{booktitle}{\emph{Proceedings of the 23rd ACM international
  conference on Multimedia}}. Acm, \bibinfo{pages}{1307--1310}.
\newblock


\bibitem[\protect\citeauthoryear{Kang, Qi, Janecek, and Banerjee}{Kang
  et~al\mbox{.}}{2015}]%
        {kang2015ecodrive}
\bibfield{author}{\bibinfo{person}{Lei Kang}, \bibinfo{person}{Bozhao Qi},
  \bibinfo{person}{Dan Janecek}, {and} \bibinfo{person}{Suman Banerjee}.}
  \bibinfo{year}{2015}\natexlab{}.
\newblock \showarticletitle{EcoDrive: A mobile sensing and control system for
  fuel efficient driving}. In \bibinfo{booktitle}{\emph{Proceedings of the 21st
  Annual International Conference on Mobile Computing and Networking}}. ACM,
  \bibinfo{pages}{358--371}.
\newblock


\bibitem[\protect\citeauthoryear{Koch, Zemel, and Salakhutdinov}{Koch
  et~al\mbox{.}}{2015}]%
        {koch2015siamese}
\bibfield{author}{\bibinfo{person}{Gregory Koch}, \bibinfo{person}{Richard
  Zemel}, {and} \bibinfo{person}{Ruslan Salakhutdinov}.}
  \bibinfo{year}{2015}\natexlab{}.
\newblock \showarticletitle{Siamese neural networks for one-shot image
  recognition}. In \bibinfo{booktitle}{\emph{ICML Deep Learning Workshop}},
  Vol.~\bibinfo{volume}{2}.
\newblock


\bibitem[\protect\citeauthoryear{Krizhevsky, Sutskever, and Hinton}{Krizhevsky
  et~al\mbox{.}}{2012}]%
        {krizhevsky2012imagenet}
\bibfield{author}{\bibinfo{person}{Alex Krizhevsky}, \bibinfo{person}{Ilya
  Sutskever}, {and} \bibinfo{person}{Geoffrey~E Hinton}.}
  \bibinfo{year}{2012}\natexlab{}.
\newblock \showarticletitle{Imagenet classification with deep convolutional
  neural networks}. In \bibinfo{booktitle}{\emph{Advances in neural information
  processing systems}}. \bibinfo{pages}{1097--1105}.
\newblock


\bibitem[\protect\citeauthoryear{Lane, Bhattacharya, Georgiev, Forlivesi, Jiao,
  Qendro, and Kawsar}{Lane et~al\mbox{.}}{2016}]%
        {lane2016deepx}
\bibfield{author}{\bibinfo{person}{Nicholas~D Lane}, \bibinfo{person}{Sourav
  Bhattacharya}, \bibinfo{person}{Petko Georgiev}, \bibinfo{person}{Claudio
  Forlivesi}, \bibinfo{person}{Lei Jiao}, \bibinfo{person}{Lorena Qendro},
  {and} \bibinfo{person}{Fahim Kawsar}.} \bibinfo{year}{2016}\natexlab{}.
\newblock \showarticletitle{Deepx: A software accelerator for low-power deep
  learning inference on mobile devices}. In
  \bibinfo{booktitle}{\emph{Proceedings of the 15th International Conference on
  Information Processing in Sensor Networks}}. IEEE Press, \bibinfo{pages}{23}.
\newblock


\bibitem[\protect\citeauthoryear{Lane, Georgiev, and Qendro}{Lane
  et~al\mbox{.}}{2015}]%
        {lane2015deepear}
\bibfield{author}{\bibinfo{person}{Nicholas~D Lane}, \bibinfo{person}{Petko
  Georgiev}, {and} \bibinfo{person}{Lorena Qendro}.}
  \bibinfo{year}{2015}\natexlab{}.
\newblock \showarticletitle{DeepEar: robust smartphone audio sensing in
  unconstrained acoustic environments using deep learning}. In
  \bibinfo{booktitle}{\emph{Proceedings of the 2015 ACM International Joint
  Conference on Pervasive and Ubiquitous Computing}}. ACM,
  \bibinfo{pages}{283--294}.
\newblock


\bibitem[\protect\citeauthoryear{Li, Sun, He, and Tan}{Li
  et~al\mbox{.}}{2017}]%
        {li2017deep}
\bibfield{author}{\bibinfo{person}{Qi Li}, \bibinfo{person}{Zhenan Sun},
  \bibinfo{person}{Ran He}, {and} \bibinfo{person}{Tieniu Tan}.}
  \bibinfo{year}{2017}\natexlab{}.
\newblock \showarticletitle{Deep supervised discrete hashing}. In
  \bibinfo{booktitle}{\emph{Advances in Neural Information Processing
  Systems}}. \bibinfo{pages}{2482--2491}.
\newblock


\bibitem[\protect\citeauthoryear{Li, An, Tian, Campbell, and Zhou}{Li
  et~al\mbox{.}}{2015}]%
        {li2015human}
\bibfield{author}{\bibinfo{person}{Tianxing Li}, \bibinfo{person}{Chuankai An},
  \bibinfo{person}{Zhao Tian}, \bibinfo{person}{Andrew~T Campbell}, {and}
  \bibinfo{person}{Xia Zhou}.} \bibinfo{year}{2015}\natexlab{}.
\newblock \showarticletitle{Human sensing using visible light communication}.
  In \bibinfo{booktitle}{\emph{Proceedings of the 21st Annual International
  Conference on Mobile Computing and Networking}}. ACM,
  \bibinfo{pages}{331--344}.
\newblock


\bibitem[\protect\citeauthoryear{LiKamWa, Hou, Gao, Polansky, and
  Zhong}{LiKamWa et~al\mbox{.}}{2016}]%
        {likamwa2016redeye}
\bibfield{author}{\bibinfo{person}{Robert LiKamWa}, \bibinfo{person}{Yunhui
  Hou}, \bibinfo{person}{Julian Gao}, \bibinfo{person}{Mia Polansky}, {and}
  \bibinfo{person}{Lin Zhong}.} \bibinfo{year}{2016}\natexlab{}.
\newblock \showarticletitle{RedEye: analog ConvNet image sensor architecture
  for continuous mobile vision}. In \bibinfo{booktitle}{\emph{ACM SIGARCH
  Computer Architecture News}}, Vol.~\bibinfo{volume}{44}. IEEE Press,
  \bibinfo{pages}{255--266}.
\newblock


\bibitem[\protect\citeauthoryear{Liu, Zhang, Han, Niu, Lai, and Xu}{Liu
  et~al\mbox{.}}{2018}]%
        {liu2018matching}
\bibfield{author}{\bibinfo{person}{Bang Liu}, \bibinfo{person}{Ting Zhang},
  \bibinfo{person}{Fred~X Han}, \bibinfo{person}{Di Niu},
  \bibinfo{person}{Kunfeng Lai}, {and} \bibinfo{person}{Yu Xu}.}
  \bibinfo{year}{2018}\natexlab{}.
\newblock \showarticletitle{Matching Natural Language Sentences with
  Hierarchical Sentence Factorization}. In
  \bibinfo{booktitle}{\emph{Proceedings of the 2018 World Wide Web Conference
  on World Wide Web}}. International World Wide Web Conferences Steering
  Committee, \bibinfo{pages}{1237--1246}.
\newblock


\bibitem[\protect\citeauthoryear{Mikolov, Karafi{\'a}t, Burget,
  {\v{C}}ernock{\`y}, and Khudanpur}{Mikolov et~al\mbox{.}}{2010}]%
        {mikolov2010recurrent}
\bibfield{author}{\bibinfo{person}{Tom{\'a}{\v{s}} Mikolov},
  \bibinfo{person}{Martin Karafi{\'a}t}, \bibinfo{person}{Luk{\'a}{\v{s}}
  Burget}, \bibinfo{person}{Jan {\v{C}}ernock{\`y}}, {and}
  \bibinfo{person}{Sanjeev Khudanpur}.} \bibinfo{year}{2010}\natexlab{}.
\newblock \showarticletitle{Recurrent neural network based language model}. In
  \bibinfo{booktitle}{\emph{Eleventh Annual Conference of the International
  Speech Communication Association}}.
\newblock


\bibitem[\protect\citeauthoryear{Miluzzo, Varshavsky, Balakrishnan, and
  Choudhury}{Miluzzo et~al\mbox{.}}{2012}]%
        {miluzzo2012tapprints}
\bibfield{author}{\bibinfo{person}{Emiliano Miluzzo},
  \bibinfo{person}{Alexander Varshavsky}, \bibinfo{person}{Suhrid
  Balakrishnan}, {and} \bibinfo{person}{Romit~Roy Choudhury}.}
  \bibinfo{year}{2012}\natexlab{}.
\newblock \showarticletitle{Tapprints: your finger taps have fingerprints}. In
  \bibinfo{booktitle}{\emph{Proceedings of the 10th international conference on
  Mobile systems, applications, and services}}. ACm, \bibinfo{pages}{323--336}.
\newblock


\bibitem[\protect\citeauthoryear{Peterek, Penhaker, Gajdo{\v{s}}, and
  Dohn{\'a}lek}{Peterek et~al\mbox{.}}{2014}]%
        {comparison14}
\bibfield{author}{\bibinfo{person}{Tom{\'a}{\v{s}} Peterek},
  \bibinfo{person}{Marek Penhaker}, \bibinfo{person}{Petr Gajdo{\v{s}}}, {and}
  \bibinfo{person}{Pavel Dohn{\'a}lek}.} \bibinfo{year}{2014}\natexlab{}.
\newblock \showarticletitle{Comparison of Classification Algorithms for
  Physical Activity Recognition}. In \bibinfo{booktitle}{\emph{Innovations in
  Bio-inspired Computing and Applications}},
  \bibfield{editor}{\bibinfo{person}{Ajith Abraham}, \bibinfo{person}{Pavel
  Kr{\"o}mer}, {and} \bibinfo{person}{V{\'a}clav Sn{\'a}{\v{s}}el}} (Eds.).
  \bibinfo{publisher}{Springer International Publishing},
  \bibinfo{address}{Cham}, \bibinfo{pages}{123--131}.
\newblock
\showISBNx{978-3-319-01781-5}


\bibitem[\protect\citeauthoryear{Radu, Lane, Bhattacharya, Mascolo, Marina, and
  Kawsar}{Radu et~al\mbox{.}}{2016}]%
        {radu2016towards}
\bibfield{author}{\bibinfo{person}{Valentin Radu}, \bibinfo{person}{Nicholas~D
  Lane}, \bibinfo{person}{Sourav Bhattacharya}, \bibinfo{person}{Cecilia
  Mascolo}, \bibinfo{person}{Mahesh~K Marina}, {and} \bibinfo{person}{Fahim
  Kawsar}.} \bibinfo{year}{2016}\natexlab{}.
\newblock \showarticletitle{Towards multimodal deep learning for activity
  recognition on mobile devices}. In \bibinfo{booktitle}{\emph{Proceedings of
  the 2016 ACM International Joint Conference on Pervasive and Ubiquitous
  Computing: Adjunct}}. ACM, \bibinfo{pages}{185--188}.
\newblock


\bibitem[\protect\citeauthoryear{Ravi, Wong, Lo, and Yang}{Ravi
  et~al\mbox{.}}{2016}]%
        {ravi2016deep}
\bibfield{author}{\bibinfo{person}{Daniele Ravi}, \bibinfo{person}{Charence
  Wong}, \bibinfo{person}{Benny Lo}, {and} \bibinfo{person}{Guang-Zhong Yang}.}
  \bibinfo{year}{2016}\natexlab{}.
\newblock \showarticletitle{Deep learning for human activity recognition: A
  resource efficient implementation on low-power devices}. In
  \bibinfo{booktitle}{\emph{Wearable and Implantable Body Sensor Networks
  (BSN), 2016 IEEE 13th International Conference on}}. IEEE,
  \bibinfo{pages}{71--76}.
\newblock


\bibitem[\protect\citeauthoryear{Ronao and Cho}{Ronao and Cho}{2014}]%
        {hiddenmc}
\bibfield{author}{\bibinfo{person}{{Charissa Ann} Ronao} {and}
  \bibinfo{person}{{Sung Bae} Cho}.} \bibinfo{year}{2014}\natexlab{}.
\newblock \showarticletitle{Human activity recognition using smartphone sensors
  with two-stage continuous hidden markov models}. In
  \bibinfo{booktitle}{\emph{2014 10th International Conference on Natural
  Computation, ICNC 2014}}. \bibinfo{publisher}{Institute of Electrical and
  Electronics Engineers Inc.}, \bibinfo{address}{United States},
  \bibinfo{pages}{681--686}.
\newblock
\urldef\tempurl%
\url{https://doi.org/10.1109/ICNC.2014.6975918}
\showDOI{\tempurl}


\bibitem[\protect\citeauthoryear{Sagha, Digumarti, Mill{\'a}n, Chavarriaga,
  Calatroni, Roggen, and Tr{\"o}ster}{Sagha et~al\mbox{.}}{2011}]%
        {sagha2011benchmarking}
\bibfield{author}{\bibinfo{person}{Hesam Sagha},
  \bibinfo{person}{Sundara~Tejaswi Digumarti}, \bibinfo{person}{Jos{\'e} del~R
  Mill{\'a}n}, \bibinfo{person}{Ricardo Chavarriaga}, \bibinfo{person}{Alberto
  Calatroni}, \bibinfo{person}{Daniel Roggen}, {and} \bibinfo{person}{Gerhard
  Tr{\"o}ster}.} \bibinfo{year}{2011}\natexlab{}.
\newblock \showarticletitle{Benchmarking classification techniques using the
  Opportunity human activity dataset}. In \bibinfo{booktitle}{\emph{Systems,
  Man, and Cybernetics (SMC), 2011 IEEE International Conference on}}. IEEE,
  \bibinfo{pages}{36--40}.
\newblock


\bibitem[\protect\citeauthoryear{Salakhutdinov and Hinton}{Salakhutdinov and
  Hinton}{2007}]%
        {salakhutdinov2007learning}
\bibfield{author}{\bibinfo{person}{Ruslan Salakhutdinov} {and}
  \bibinfo{person}{Geoff Hinton}.} \bibinfo{year}{2007}\natexlab{}.
\newblock \showarticletitle{Learning a nonlinear embedding by preserving class
  neighbourhood structure}. In \bibinfo{booktitle}{\emph{Artificial
  Intelligence and Statistics}}. \bibinfo{pages}{412--419}.
\newblock


\bibitem[\protect\citeauthoryear{S{\'a}nchez and Perronnin}{S{\'a}nchez and
  Perronnin}{2011}]%
        {sanchez2011high}
\bibfield{author}{\bibinfo{person}{Jorge S{\'a}nchez} {and}
  \bibinfo{person}{Florent Perronnin}.} \bibinfo{year}{2011}\natexlab{}.
\newblock \showarticletitle{High-dimensional signature compression for
  large-scale image classification}. In \bibinfo{booktitle}{\emph{Computer
  Vision and Pattern Recognition (CVPR), 2011 IEEE Conference on}}. IEEE,
  \bibinfo{pages}{1665--1672}.
\newblock


\bibitem[\protect\citeauthoryear{Shen, Shen, Liu, and Tao~Shen}{Shen
  et~al\mbox{.}}{2015}]%
        {shen2015supervised}
\bibfield{author}{\bibinfo{person}{Fumin Shen}, \bibinfo{person}{Chunhua Shen},
  \bibinfo{person}{Wei Liu}, {and} \bibinfo{person}{Heng Tao~Shen}.}
  \bibinfo{year}{2015}\natexlab{}.
\newblock \showarticletitle{Supervised discrete hashing}. In
  \bibinfo{booktitle}{\emph{CVPR}}. \bibinfo{pages}{37--45}.
\newblock


\bibitem[\protect\citeauthoryear{Stisen, Blunck, Bhattacharya, Prentow,
  Kj{\ae}rgaard, Dey, Sonne, and Jensen}{Stisen et~al\mbox{.}}{2015}]%
        {stisen2015smart}
\bibfield{author}{\bibinfo{person}{Allan Stisen}, \bibinfo{person}{Henrik
  Blunck}, \bibinfo{person}{Sourav Bhattacharya}, \bibinfo{person}{Thor~Siiger
  Prentow}, \bibinfo{person}{Mikkel~Baun Kj{\ae}rgaard}, \bibinfo{person}{Anind
  Dey}, \bibinfo{person}{Tobias Sonne}, {and} \bibinfo{person}{Mads~M{\o}ller
  Jensen}.} \bibinfo{year}{2015}\natexlab{}.
\newblock \showarticletitle{Smart devices are different: Assessing and
  mitigatingmobile sensing heterogeneities for activity recognition}. In
  \bibinfo{booktitle}{\emph{Proceedings of the 13th ACM Conference on Embedded
  Networked Sensor Systems}}. ACM, \bibinfo{pages}{127--140}.
\newblock


\bibitem[\protect\citeauthoryear{Vedaldi and Zisserman}{Vedaldi and
  Zisserman}{2012}]%
        {vedaldi2012sparse}
\bibfield{author}{\bibinfo{person}{Andrea Vedaldi} {and}
  \bibinfo{person}{Andrew Zisserman}.} \bibinfo{year}{2012}\natexlab{}.
\newblock \showarticletitle{Sparse kernel approximations for efficient
  classification and detection}. In \bibinfo{booktitle}{\emph{Computer Vision
  and Pattern Recognition (CVPR), 2012 IEEE Conference on}}. IEEE,
  \bibinfo{pages}{2320--2327}.
\newblock


\bibitem[\protect\citeauthoryear{Vinyals, Blundell, Lillicrap, Wierstra,
  et~al\mbox{.}}{Vinyals et~al\mbox{.}}{2016}]%
        {vinyals2016matching}
\bibfield{author}{\bibinfo{person}{Oriol Vinyals}, \bibinfo{person}{Charles
  Blundell}, \bibinfo{person}{Timothy Lillicrap}, \bibinfo{person}{Daan
  Wierstra}, {et~al\mbox{.}}} \bibinfo{year}{2016}\natexlab{}.
\newblock \showarticletitle{Matching networks for one shot learning}. In
  \bibinfo{booktitle}{\emph{Advances in neural information processing
  systems}}. \bibinfo{pages}{3630--3638}.
\newblock


\bibitem[\protect\citeauthoryear{Wan, Qi, Xu, Tong, and Gu}{Wan
  et~al\mbox{.}}{2020}]%
        {wan2020deep}
\bibfield{author}{\bibinfo{person}{Shaohua Wan}, \bibinfo{person}{Lianyong Qi},
  \bibinfo{person}{Xiaolong Xu}, \bibinfo{person}{Chao Tong}, {and}
  \bibinfo{person}{Zonghua Gu}.} \bibinfo{year}{2020}\natexlab{}.
\newblock \showarticletitle{Deep learning models for real-time human activity
  recognition with smartphones}.
\newblock \bibinfo{journal}{\emph{Mobile Networks and Applications}}
  \bibinfo{volume}{25}, \bibinfo{number}{2} (\bibinfo{year}{2020}),
  \bibinfo{pages}{743--755}.
\newblock


\bibitem[\protect\citeauthoryear{Wang, Guo, Wang, Chen, and Liu}{Wang
  et~al\mbox{.}}{2016}]%
        {wang2016friend}
\bibfield{author}{\bibinfo{person}{Chen Wang}, \bibinfo{person}{Xiaonan Guo},
  \bibinfo{person}{Yan Wang}, \bibinfo{person}{Yingying Chen}, {and}
  \bibinfo{person}{Bo Liu}.} \bibinfo{year}{2016}\natexlab{}.
\newblock \showarticletitle{Friend or foe?: Your wearable devices reveal your
  personal pin}. In \bibinfo{booktitle}{\emph{Proceedings of the 11th ACM on
  Asia Conference on Computer and Communications Security}}. ACM,
  \bibinfo{pages}{189--200}.
\newblock


\bibitem[\protect\citeauthoryear{Wang, Chen, Hao, Peng, and Hu}{Wang
  et~al\mbox{.}}{2018a}]%
        {wang2018deep}
\bibfield{author}{\bibinfo{person}{Jindong Wang}, \bibinfo{person}{Yiqiang
  Chen}, \bibinfo{person}{Shuji Hao}, \bibinfo{person}{Xiaohui Peng}, {and}
  \bibinfo{person}{Lisha Hu}.} \bibinfo{year}{2018}\natexlab{a}.
\newblock \showarticletitle{Deep learning for sensor-based activity
  recognition: A survey}.
\newblock \bibinfo{journal}{\emph{Pattern Recognition Letters}}
  (\bibinfo{year}{2018}).
\newblock


\bibitem[\protect\citeauthoryear{Wang, Zhang, Sebe, Shen, et~al\mbox{.}}{Wang
  et~al\mbox{.}}{2018b}]%
        {wang2018survey}
\bibfield{author}{\bibinfo{person}{Jingdong Wang}, \bibinfo{person}{Ting
  Zhang}, \bibinfo{person}{Nicu Sebe}, \bibinfo{person}{Heng~Tao Shen},
  {et~al\mbox{.}}} \bibinfo{year}{2018}\natexlab{b}.
\newblock \showarticletitle{A survey on learning to hash}.
\newblock \bibinfo{journal}{\emph{IEEE Transactions on Pattern Analysis and
  Machine Intelligence}} \bibinfo{volume}{40}, \bibinfo{number}{4}
  (\bibinfo{year}{2018}), \bibinfo{pages}{769--790}.
\newblock


\bibitem[\protect\citeauthoryear{Yao, Hu, Zhao, Zhang, and Abdelzaher}{Yao
  et~al\mbox{.}}{2017}]%
        {yao2017deepsense}
\bibfield{author}{\bibinfo{person}{Shuochao Yao}, \bibinfo{person}{Shaohan Hu},
  \bibinfo{person}{Yiran Zhao}, \bibinfo{person}{Aston Zhang}, {and}
  \bibinfo{person}{Tarek Abdelzaher}.} \bibinfo{year}{2017}\natexlab{}.
\newblock \showarticletitle{Deepsense: A unified deep learning framework for
  time-series mobile sensing data processing}. In
  \bibinfo{booktitle}{\emph{Proceedings of the 26th International Conference on
  World Wide Web}}. International World Wide Web Conferences Steering
  Committee, \bibinfo{pages}{351--360}.
\newblock


\bibitem[\protect\citeauthoryear{yscacaca}{yscacaca}{2019}]%
        {deepsense_pro}
\bibfield{author}{\bibinfo{person}{yscacaca}.} \bibinfo{year}{2019}\natexlab{}.
\newblock \bibinfo{booktitle}{\emph{HHAR-Data-Process}}.
\newblock
\newblock
\shownote{\url{https://github.com/yscacaca/HHAR-Data-Process}.}


\bibitem[\protect\citeauthoryear{Zhang and Sawchuk}{Zhang and Sawchuk}{2012}]%
        {zhang2012usc}
\bibfield{author}{\bibinfo{person}{Mi Zhang} {and} \bibinfo{person}{Alexander~A
  Sawchuk}.} \bibinfo{year}{2012}\natexlab{}.
\newblock \showarticletitle{USC-HAD: a daily activity dataset for ubiquitous
  activity recognition using wearable sensors}. In
  \bibinfo{booktitle}{\emph{Proceedings of the 2012 ACM Conference on
  Ubiquitous Computing}}. ACM, \bibinfo{pages}{1036--1043}.
\newblock


\bibitem[\protect\citeauthoryear{Zhu, Wang, Zhang, and Xu}{Zhu
  et~al\mbox{.}}{2014}]%
        {zhu2014human}
\bibfield{author}{\bibinfo{person}{Yangda Zhu}, \bibinfo{person}{Changhai
  Wang}, \bibinfo{person}{Jianzhong Zhang}, {and} \bibinfo{person}{Jingdong
  Xu}.} \bibinfo{year}{2014}\natexlab{}.
\newblock \showarticletitle{Human activity recognition based on similarity}. In
  \bibinfo{booktitle}{\emph{2014 IEEE 17th International Conference on
  Computational Science and Engineering}}. IEEE, \bibinfo{pages}{1382--1387}.
\newblock


\end{thebibliography}
\end{document}